\documentclass{article}
\usepackage{algorithm}
\usepackage{algorithmic}
\PassOptionsToPackage{numbers, compress}{natbib}

\usepackage[main, final]{neurips_2026}

\usepackage[utf8]{inputenc} 
\usepackage[T1]{fontenc}    
\usepackage{hyperref}       
\usepackage{url}            
\usepackage{booktabs}       
\usepackage{amsfonts}       
\usepackage{nicefrac}       
\usepackage{microtype}      
\usepackage{xcolor}         

\usepackage{multirow}
\usepackage{colortbl}
\usepackage{subfigure}
\usepackage{amsmath}
\usepackage{amssymb}
\usepackage{wrapfig}
\usepackage{enumitem}
\usepackage{amsthm}
\usepackage{graphicx}
\usepackage{array}
\usepackage{framed}
\definecolor{promptbg}{RGB}{246,246,250}
\definecolor{prompttitle}{RGB}{40,70,135}
\colorlet{shadecolor}{promptbg}

\title{HyperSkill: Training-Free Omnimodal GRPO via Hypergraph-Indexed Skill-Library Evolution}

\author{
{\bf Haoran Luo\textsuperscript{\rm 1}\textsuperscript{$\dagger$}, Shangkai Lin\textsuperscript{\rm 1}, Jinyang Wu\textsuperscript{\rm 2}, Xinliang Zhou\textsuperscript{\rm 3}\textsuperscript{$*$}, Luu Anh Tuan\textsuperscript{\rm 1,4}\textsuperscript{$*$}} \\
         \textsuperscript{1}Nanyang Technological University\ \ \textsuperscript{2}Tsinghua University\ \ \textsuperscript{3}Stanford University\ \ \textsuperscript{4}VinUniversity \\ 
         \texttt{haoran.luo@ntu.edu.sg, xinliang.zhou@stanford.edu, anhtuan.luu@ntu.edu.sg}
}

\newsavebox{\tabmainbox}\newsavebox{\tababbox}
\begin{document}

\maketitle
{\renewcommand{\thefootnote}{$\dagger$}\footnotetext{Project Lead.\quad\textsuperscript{$*$}Corresponding Authors.\quad\textsuperscript{1}\url{https://github.com/LHRLAB/HyperSkill}}}

\vspace{-2mm}

\begin{abstract}
Reinforcement learning sharpens LLM reasoning, but parameter updates are expensive, forgetful, and opaque; training-free libraries swap gradients for a textual store but still use trajectory-level, correlation-based admission and flat structure. To address both, we propose \textbf{HyperSkill}, a \emph{Training-Free Omnimodal GRPO} framework that keeps the backbone frozen and replaces gradient updates with auditable edits to a \emph{Hypergraph-Indexed Skill-Library}, where skills are advantage-bearing nodes and compositions are hyperedges. It has three components: training-time distillation admits counterfactually checked skills and learns a protocol skill from failures; gated hypergraph retrieval injects advantage-ranked skills where a per-dataset gate admits them; and an execution protocol makes the frozen backbone commit to an answer. On 11 text, vision-language, and omnimodal benchmarks, HyperSkill is best or tied on every dataset and beats the best baseline by 13.0 average F1 points while keeping the backbone frozen. Our software and data are publicly available\textsuperscript{1}.
\end{abstract}

\section{Introduction}
\label{Section1}

\begin{wrapfigure}{r}{0.54\textwidth}
\vspace{-4.5mm}
\centering
\includegraphics[width=\linewidth]{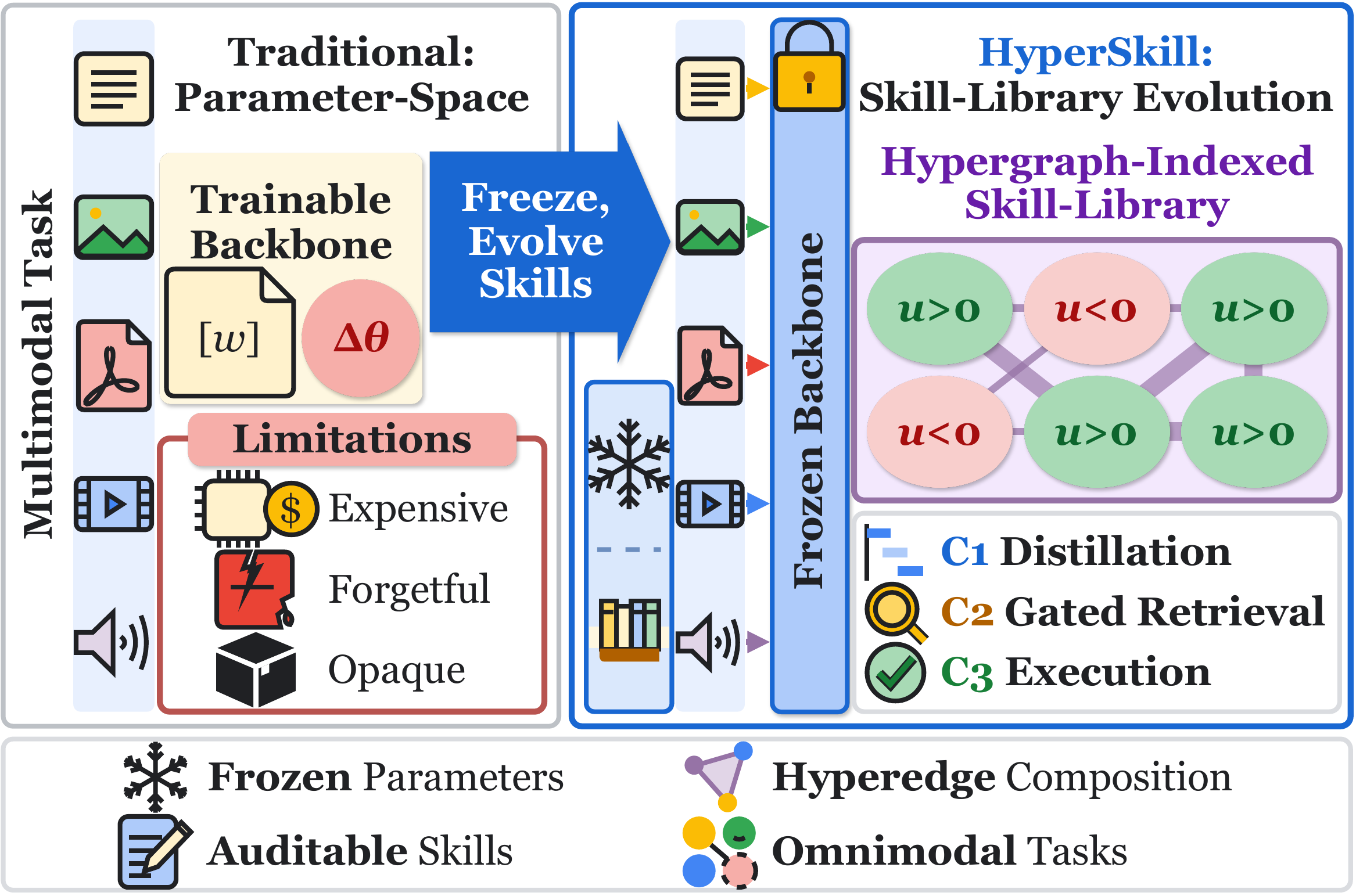}
\vspace{-5mm}
\caption{\parfillskip=0pt\relax Parameter-space GRPO is costly, forgetful \protect\linebreak and opaque; HyperSkill instead keeps the backbone \protect\linebreak frozen and evolves a hypergraph-indexed skill library.}\label{F1}
\vspace{-3mm}
\end{wrapfigure}

Reinforcement learning is now the main way to improve the reasoning of large language models (LLMs), with group-relative policy optimisation (GRPO)~\citep{GRPO-DeepSeekMath,DAPO,GSPO} used to train systems in math, search, and tool use~\citep{DeepSeek-R1,Qwen3,Search-R1,ReTool}. However, it is still hard to keep improving an agent as it moves to omnimodal tasks such as web search, long-document reasoning, and audio-video question answering~\citep{WebWalker,Qwen3-Omni,OmniRAG-Agent,OmniGAIA}, without a per-domain cost of updating its backbone.

Existing methods follow two main lines. The mainstream parameter-space recipe applies gradient updates to the backbone, yielding capable but costly systems~\citep{DeepSeek-R1,Qwen3,Search-R1,ReTool}. A growing training-free line keeps the backbone frozen and grows an external textual experience store across rollouts, retrieved in-context at inference~\citep{TrainingFreeGRPO,EvoSkill,SkillCraft,AutoSkill,Voyager,JARVIS-1}.

\begin{figure}[t]
\centering
\includegraphics[width=\linewidth]{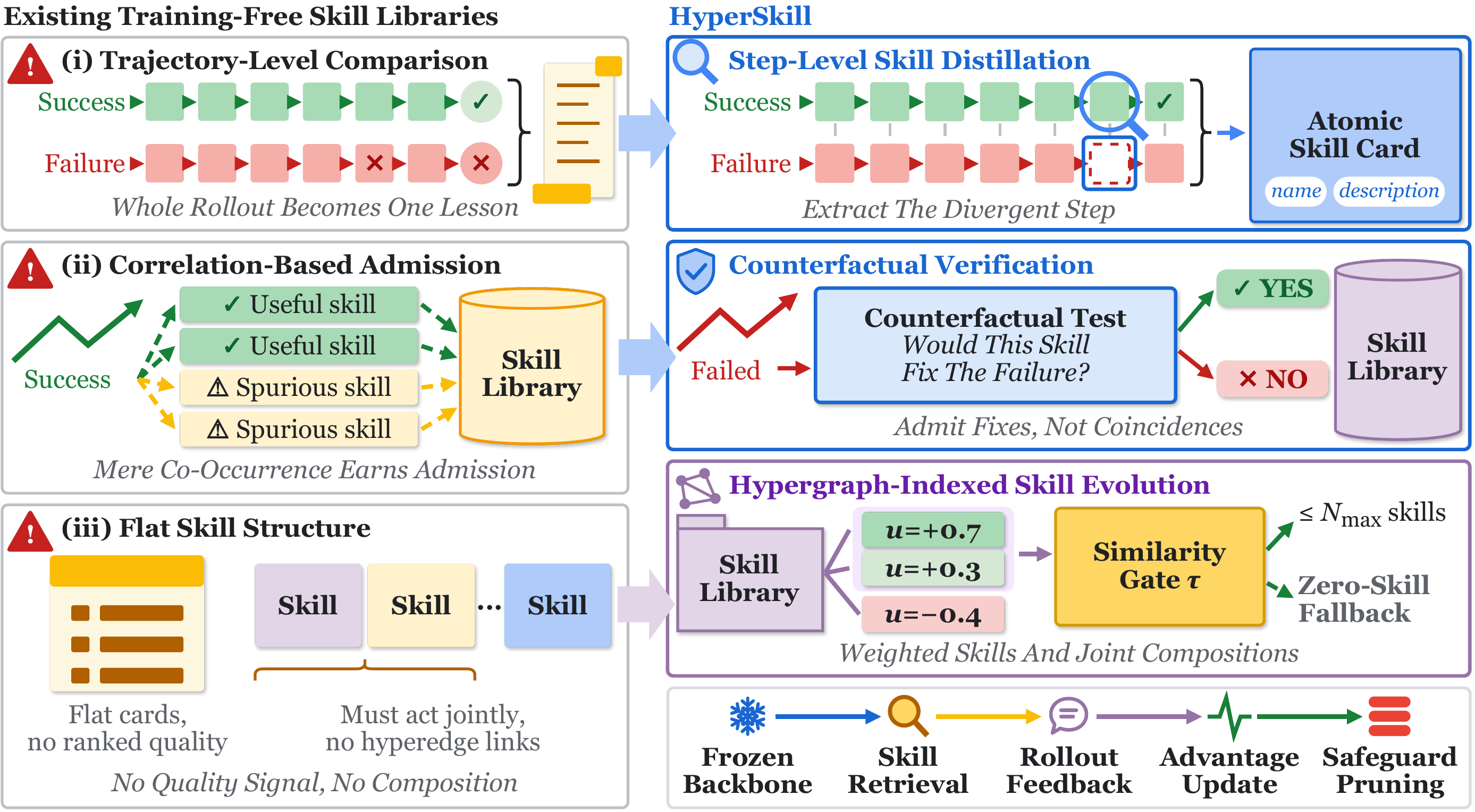} 
\caption{Three failure modes (i)--(iii) of training-free libraries (trajectory-level, correlation-based, flat skills) vs.\ HyperSkill's step distillation, counterfactual verification and hypergraph evolution.}\label{F2}
\vspace{-2mm}
\end{figure}

Both lines have limits. On the parameter side (Figure~\ref{F1}), gradient updates remain expensive at frontier scale~\citep{DeepSeek-R1}, every new task risks catastrophic forgetting~\citep{LLMsurvey}, and improvement is buried in opaque weight deltas with no auditable trace of \emph{which} patterns the agent has internalised. On the training-free side, three failure modes recur (Figure~\ref{F2}): (i) comparison is at the trajectory level~\citep{TrainingFreeGRPO,EvoSkill}, so every divergent step is fused into one lesson; (ii) candidates are admitted by co-occurrence with successful rollouts~\citep{TrainingFreeGRPO,SkillCraft,AutoSkill}, so the library keeps patterns that track reward but do not fix the failure; (iii) skills are stored as dictionaries or uniform-quality lists, which cannot express that \emph{several skills must act jointly} or give a per-skill quality signal. Hypergraph retrieval~\citep{HyperGraphRAG,HyperRAG,HyperMem} links several entities in one hyperedge, but it targets static factual knowledge bases rather than self-evolving skill libraries.

To close all three gaps, we propose HyperSkill, a \emph{Training-Free Omnimodal GRPO} framework whose policy is an external \emph{Hypergraph-Indexed Skill-Library}: atomic skills are advantage-bearing nodes, joint compositions are hyperedges, and the library is the optimisation target. \textbf{C1} (training-time distillation) handles (i) and (ii): it admits a step-level candidate only after a counterfactual test (``would the failed trajectory have succeeded had the agent known this skill?'') and learns a protocol skill from failures. \textbf{C2} (gated hypergraph retrieval) handles (iii): an advantage-weighted retriever~\citep{RAG-original,RAGSurvey,LightRAG} ranks hyperedge-linked skills~\citep{HyperGraphRAG,HyperRAG,Graph-R1} behind a gate with zero-skill fallback. \textbf{C3} (execution protocol) makes the frozen backbone call its tools correctly and then commit to a final answer.

We validate this design across 11 text, vision-language, and omnimodal benchmarks~\citep{WebWalker,InfoSeek,MMLongBench-Doc,WorldSense}: HyperSkill is best or tied on all 11 without updating a parameter, and lifts the average F1 from $29.0$ for the strongest baseline, XSkill~\citep{XSkill}, to $42.0$ ($+13.0$, $44.9\%$ relative). Each component matters: removing C1, C2 or C3 alone lowers the average F1 by $2.5$, $4.2$ and $10.4$ points, respectively.

\section{Related Work}
\label{sec:related}
\textbf{Omnimodal LLMs and reinforcement learning.} Modern backbones fuse text, image, video, and audio into a single policy~\citep{CLIP,Llava,Whisper,GPT4,Qwen3-Omni,OmniRAG-Agent,OmniGAIA}, and group-relative policy optimisation~\citep{GRPO-DeepSeekMath,DAPO,GSPO} trains reasoners~\citep{DeepSeek-R1,GPT-o1} and search and tool agents~\citep{Search-R1,ReTool,Graph-R1}, while reflection-style agents~\citep{ReAct,Reflexion,SelfRefine} act through prompting on web and OS benchmarks~\citep{WebArena,WebWalker,OSWorld}. The RL-trained ones still pay for backbone gradient updates and inherit the forgetting and opacity problems motivating this work. \emph{Training-Free Omnimodal GRPO} keeps the backbone frozen and redirects a normalised reward advantage from the gradient onto an external \emph{Hypergraph-Indexed Skill-Library} that acts as the policy, so each improvement is a readable library edit rather than an opaque weight delta.

\textbf{Training-free experience self-evolution.} TrainingFreeGRPO~\citep{TrainingFreeGRPO} replaces the gradient with an LLM-curated experience pool, skill-centric agents~\citep{EvoSkill,SkillCraft,AutoSkill,Voyager,JARVIS-1} grow flat dictionaries or frontier sets, and memory~\citep{MemGPT,Mem0,MemoRAG} with graph and hypergraph RAG~\citep{GraphRAG,LightRAG,HyperGraphRAG,HyperRAG} supplies the retrievable side. Most of them compare \emph{whole trajectories} and keep flat skills with no per-item quality for ranking; HyperSkill instead distils individual steps and ranks skills by advantage in a hypergraph.

\clearpage

\begin{table}[t]
\centering
\caption{Comparison of training-free skill methods. TrainingFreeGRPO: token-prior experience pool; EvoSkill: Pareto frontier of skill folders; SkillCraft/AutoSkill: composed-tool code or interaction-trace cards; Memento-Skills/XSkill: markdown router memory or dual experience-skill stream; HyperSkill (Ours): weighted hypergraph with counterfactual distillation and gated RRF.}
\label{tab:comparison}
\vspace{6pt}
\fontsize{8.5pt}{9.7pt}\selectfont
\centering
\setlength{\tabcolsep}{2pt}
{%
\begin{tabular}{p{79pt}|p{149pt}|p{156.7pt}}
\toprule
\textbf{Method} & \textbf{Skill Organisation} & \textbf{Retrieval \& Update} \\
\midrule
TrainingFreeGRPO~\citep{TrainingFreeGRPO} & $\mathcal{L}=\{x_i\}_{i=1}^{|\mathcal{L}|}$\newline Experiential token-prior pool from rollouts. & $\mathcal{R}(q)=\mathrm{Top\text{-}}k_{x_i}\,\mathrm{cos}(f(q),f(x_i))$\newline LLM CRUD via group-relative semantic adv. \\
EvoSkill~\citep{EvoSkill}                 & $\mathcal{L}_{\text{frt}}\!=\!\mathrm{Pareto}\{\mathrm{folder}_i\}$\newline Pareto frontier of structured skill folders. & $\mathcal{R}(q)=\mathrm{Top\text{-}}k_{s_i\in\mathcal{L}_{\text{frt}}}\,\mathrm{cos}$\newline Failure-driven propose, holdout-validate. \\
\midrule
SkillCraft~\citep{SkillCraft}             & $\mathcal{L}=\{\,\mathtt{def}\ s_i(\cdot)\,\}$\newline Auto-composed tool-Skill code cache. & $\mathcal{R}(q)=\mathrm{Top\text{-}}k_{s_i}\,\mathrm{cos}$\newline Cache-hit reuse, no quality signal. \\
AutoSkill~\citep{AutoSkill}               & $\mathcal{L}=\{\mathrm{card}_i\}$\newline Standardised cards from interaction traces. & $\mathcal{R}(q)=\mathrm{Top\text{-}}k_{s_i}\,\mathrm{cos}$\newline Dynamic injection, self-evolving lifelong. \\
\midrule
Memento-Skills~\citep{Memento-Skills}     & $\mathcal{L}=\{m_i^{\text{md}}\}$\newline Markdown skills as stateful memory. & $\mathcal{R}(q)=\mathrm{Router}(q,\mathcal{L})$\newline Read/write reflective skill router. \\
XSkill~\citep{XSkill}                     & $\{\mathcal{X}_{\mathrm{exp}},\mathcal{L}_{\mathrm{skill}}\}$\newline Dual experience and skill streams. & $\mathcal{R}(q,I)=\mathrm{Top\text{-}}k_{\,(\mathcal{X},\mathcal{L})}\,\mathrm{cos}$\newline Vision-grounded retrieve $+$ cross-critique. \\
\midrule
\textbf{HyperSkill (Ours)}       & $\mathcal{H}\!=\!(\mathcal{V},\mathcal{E},w),\ u_i\!\in\![-1,1]$\newline Weighted hypergraph with hyperedges. & $\mathcal{R}(q)=\mathrm{RRF}_{u_i}(\mathcal{V},\mathcal{E})\!\circ\!\mathrm{Gate}_{\tau,N_{\max}}$\newline Counterfactual admit + 0-skill fallback. \\
\bottomrule
\end{tabular}
}
\vspace{-1.0pt}
\end{table}

\section{Methodology: HyperSkill}
\label{sec:method}

\begin{figure*}[t]
\centering
\includegraphics[width=\linewidth]{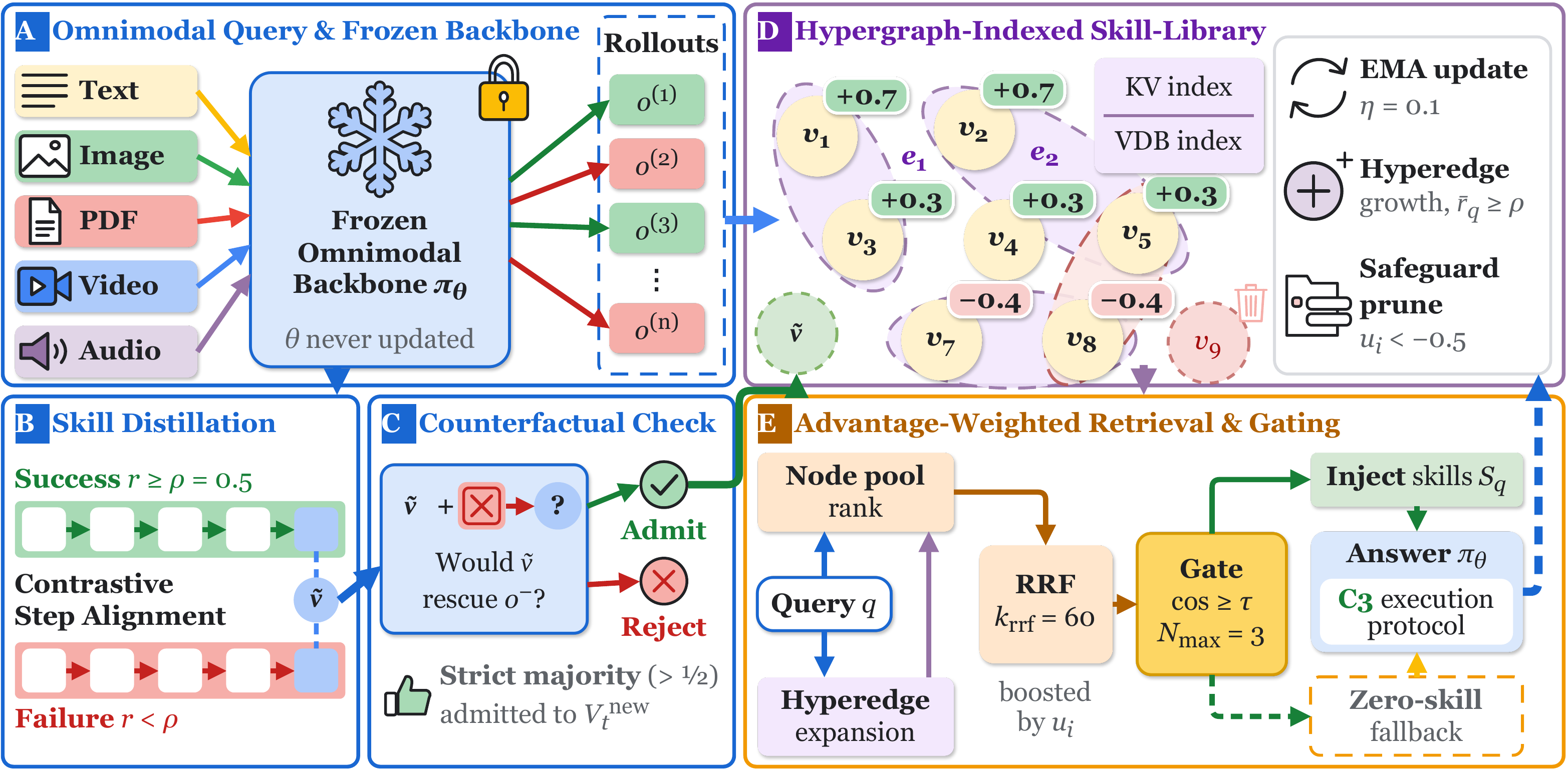} 
\caption{Methodology of HyperSkill: a frozen backbone consults a weighted skill hypergraph built by C1 (panels B--C) and evolved and retrieved by C2 (panels D--E); C3 (\S\ref{sec:exec}) governs execution.}
\label{F3}
\end{figure*}

\noindent\textbf{Problem Formulation.} Unlike the methods of Table~\ref{tab:comparison}, HyperSkill couples a per-skill quality signal with explicit multi-skill compositions in a weighted skill hypergraph. Formally, let $\pi_{\theta}$ be a frozen omnimodal backbone whose parameters $\theta$ are never updated, and let $\mathcal{D}_{\mathrm{train}}$ and $\mathcal{D}_{\mathrm{test}}$ be train and test sets of triples $(q,a^{*},m)$, in which each query $q$ carries a gold answer $a^{*}$ and a modality tag $m\!\in\!\mathcal{M}\!=\!\{\text{text},\text{image},\text{pdf},\text{video},\text{audio}\}$. We seek a weighted skill hypergraph $\mathcal{H}\!=\!(\mathcal{V},\mathcal{E},w)$ in which every node $v_{i}\!\in\!\mathcal{V}$ carries a name, a description and an advantage $u_{i}\!\in\![-1,1]$, and every hyperedge $e_{j}\!\in\!\mathcal{E}$ groups $|e_{j}|\!\geq\!2$ nodes into a joint composition with weight $w(e_{j})$. We also seek a retriever $\mathcal{R}\!:\!(q,m,\mathcal{H})\!\to\!\mathcal{S}_{q}\!\subset\!\mathcal{V}$ under which the skill-augmented policy improves on its skill-free version, while a hypergraph update $\mathcal{U}$ replaces the gradient step in every round $t$:
\begin{equation}
\bigl(\mathcal{H}^{\star},\,\mathcal{R}^{\star}\bigr)\;=\;\arg\!\max_{\mathcal{H},\,\mathcal{R}}\;\mathbb{E}_{(q,a^{*},m)\sim\mathcal{D}}\!\Bigl[\,r\bigl(\,\pi_{\theta}(\cdot\!\mid\!q,m,\mathcal{R}(q,m,\mathcal{H}))\,;\,a^{*}\,\bigr)\,\Bigr]\,,\quad\mathcal{H}_{t+1}\!=\!\mathcal{U}(\mathcal{H}_{t};\mathcal{B}_{t})\,,
\end{equation}
where $\mathcal{D}$ is the task distribution ($\mathcal{D}_{\mathrm{train}}$ in training, $\mathcal{D}_{\mathrm{test}}$ at evaluation), $\mathcal{B}_{t}$ the round-$t$ training batch and $r$ the token-level $F_{1}$ reward against $a^{*}$. Unlike GRPO, the optimisation variable is the pair $(\mathcal{H},\mathcal{R})$ rather than $\theta$: gradient updates are replaced by an iterative, auditable rewrite of the library. C1 (\S\ref{sec:distill}) handles gaps (i) and (ii), C2 (\S\ref{sec:hypergraph}) handles gap (iii), and C3 (\S\ref{sec:exec}) fixes execution failures of the frozen backbone; Figure~\ref{F3} gives an overview of how C1 and C2 build, evolve and query the library.

\subsection{C1: Training-Time Skill Distillation}
\label{sec:distill}

\noindent To address (i) and (ii), we move admission from \emph{trajectories} to \emph{steps} and route every candidate through a counterfactual probe before insertion, so that a skill enters $\mathcal{V}$ only if the frozen backbone judges it to salvage a failure of its own question rather than merely co-occurring with a success.

\noindent\textbf{Contrastive Step Alignment.} For every question $q\!\in\!\mathcal{B}_{t}$ we draw a group of $n$ skill-augmented rollouts $\{o^{(\ell)}\}_{\ell=1}^{n}\!\sim\!\pi_{\theta}(\cdot\!\mid\!q,m,\mathcal{R}(q,m,\mathcal{H}_{t}))$ under the frozen backbone, score each by the token-level $F_{1}$ reward, and split the group at threshold $\rho\!=\!0.5$ into a success bucket $\mathcal{O}_{q}^{+}$ and a failure bucket $\mathcal{O}_{q}^{-}$. A regex decomposer $\Psi$ (zero LLM cost) slices every rollout into a step sequence $\Psi(o)\!=\!(s_{1},\dots,s_{T_{o}})$, after which an LLM aligner $\Phi$ matches steps that solve the same sub-goal across the two buckets and harvests each divergent atomic skill that successes show but failures miss:
\begin{equation}
\widetilde{\mathcal{V}}_{q}\;=\;\Phi\!\Bigl(\,\bigl\{\,\Psi(o^{(\ell)})\!:\!r(o^{(\ell)};a^{*})\!\geq\!\rho\,\bigr\}_{\ell=1}^{n}\,,\;\;\bigl\{\,\Psi(o^{(\ell)})\!:\!r(o^{(\ell)};a^{*})\!<\!\rho\,\bigr\}_{\ell=1}^{n}\,\Bigr),
\end{equation}
where each candidate $\widetilde{v}\!\in\!\widetilde{\mathcal{V}}_{q}$ is a (name, description) pair, and the aligner may also name a \emph{recipe} $\widetilde{e}\!\subset\!\widetilde{\mathcal{V}}_{q}$ of candidates that the successes used jointly, which becomes a hyperedge once at least two of its members are admitted. The contrast is thus localised to the divergence point of the two buckets, so noisy steps of successful rollouts are less likely to enter the harvested skill or recipe.

\noindent\textbf{Counterfactual Verification.} Step alignment alone still admits skills that merely correlate with success; we therefore pair each candidate with a list $\mathcal{P}_{q}$ of at most $K$ failure probes $o^{-}\!\in\!\mathcal{O}_{q}^{-}$ of its own question and ask the same frozen $\pi_{\theta}$ a binary counterfactual probe~\citep{CausalRAG,SelfRefine,Reflexion}: ``had the agent known $\widetilde{v}$, would $o^{-}$ likely have produced $a^{*}$?''. Only when a strict majority of the probes return YES is the candidate admitted into the round-$t$ skill increment, and a deterministic identifier $\mathrm{sid}(\widetilde{v})\!=\!\mathrm{md5}(\mathrm{name}(\widetilde{v}))$ deduplicates it against the current library $\mathcal{V}_{t}$ by skill name:
\begin{equation}
\mathcal{V}_{t}^{\mathrm{new}}\;=\;\Bigl\{\;\widetilde{v}\!\in\!\widetilde{\mathcal{V}}_{q}\;:\;\frac{1}{|\mathcal{P}_{q}|}\!\sum_{o^{-}\in\mathcal{P}_{q}}\!\mathbf{1}\!\bigl[\,\pi_{\theta}\!\bigl(\,\cdot\!\mid\!\widetilde{v},\,q,\,a^{*},\,o^{-}\bigr)\!=\!\mathrm{YES}\,\bigr]\;>\;\tfrac{1}{2}\;\;\Bigr\}\,,\qquad |\mathcal{P}_{q}|\!\leq\!K\!=\!2.
\end{equation}
The admitted increment $\mathcal{V}_{t}^{\mathrm{new}}$ and its recipes $\mathcal{E}_{t}^{\mathrm{rec}}$ feed the hypergraph evolution of C2 (\S\ref{sec:hypergraph}).

\noindent\textbf{Protocol Skill and Answer Form.} A frozen backbone often breaks the tool or answer format, not the task. C1 therefore sorts every training rollout that ends without an answer into four failure types: malformed tool call, call beyond the budget, call without tools, and no \texttt{<answer>} tag. Each tool-related type seen in at least $2$ rollouts adds fixed instruction lines, filled with the dataset's tool tags and budget, to a \emph{protocol skill}. Its last line fixes the answer form: the integer without \texttt{\textbackslash boxed\{\}} on math, else a short answer or, per dataset, a form derived from the training gold answers (option letter, phrase or sentence, plus an abstention line). The protocol skill is not a node of $\mathcal{H}$ (App.~\ref{app:prompts}).

\textbf{Proposition 1.} \textit{Counterfactual distillation favours failure-rescuing skills over co-occurring ones.}
\vspace{-2mm}
\begin{proof}
We provide experimental results in \S\ref{sec:rq2} and \S\ref{sec:rq3} and proofs in Appendix~\ref{app:proof1}.
\end{proof}

\subsection{C2: Hypergraph Evolution and Gated Retrieval}
\label{sec:hypergraph}\label{sec:gating}

\noindent To address (iii), the survivors of C1 form a weighted hypergraph $\mathcal{H}\!=\!(\mathcal{V},\mathcal{E},w)$. During training, rollout rewards are recycled, with no extra LLM call, into a per-node advantage and per-edge weights; at test time, with the library frozen, C2 ranks, gates and injects skills for every question.

\noindent\textbf{Hypergraph Index and Advantage EMA Update.} Each admitted skill $v_{i}$ is embedded by a frozen text encoder $f$ and stored with its advantage $u_{i}$ in a dual key-value/vector-database (KV/VDB) index; admitted recipes and skill sets retrieved together for high-reward questions become hyperedges $e_{j}\!\subset\!\mathcal{V}$ with $|e_{j}|\!\geq\!2$. After every batch, the batch-normalised mean reward of each question $q$ is broadcast to its retrieved skills $\mathcal{S}_{q}$, and an exponential moving average with rate $\eta$ updates their advantages $u_{i}$:
\begin{equation}
\hat{A}_{q}\;=\;\tanh\!\Bigl(\,\tfrac{\bar{r}_{q}\,-\,\bar{r}_{\mathcal{B}_{t}}}{\sigma_{r,\mathcal{B}_{t}}\,+\,\epsilon}\,\Bigr)\,,\quad u_{i}\;\leftarrow\;\mathrm{clip}_{[-1,1]}\!\bigl(\,(1\!-\!\eta)\,u_{i}\,+\,\eta\,\hat{A}_{q}\,\bigr)\;\;\forall\,v_{i}\!\in\!\mathcal{S}_{q}\,,
\end{equation}
where $\bar{r}_{q}$ is the mean reward of the $n$ rollouts of $q$ and $(\bar{r}_{\mathcal{B}_{t}},\sigma_{r,\mathcal{B}_{t}})$ are statistics over all rollouts of the batch, with $(\eta,\epsilon)\!=\!(0.1,10^{-8})$. Hyperedges grow whenever a high-reward question retrieves two or more skills jointly, and a periodic pruning step removes every node whose EMA advantage falls below a threshold $u_{\min}$; together, the node and hyperedge updates of round $t$ read:
\begin{equation}
\mathcal{V}_{t+1}\;=\;\bigl\{\,v_{i}\!\in\!\mathcal{V}_{t}\!\cup\!\mathcal{V}_{t}^{\mathrm{new}}\,:\,u_{i}\!\geq\!u_{\min}\,\bigr\}\,,\qquad \mathcal{E}_{t+1}\;=\;\mathcal{E}_{t}\,\cup\,\mathcal{E}_{t}^{\mathrm{rec}}\,\cup\,\bigl\{\,\mathcal{S}_{q}\,:\,\bar{r}_{q}\!\geq\!\rho,\;|\mathcal{S}_{q}|\!\geq\!2\,\bigr\}\,,
\end{equation}
with $u_{\min}\!=\!-0.5$ and $\mathcal{E}_{t}^{\mathrm{rec}}$ the round-$t$ recipes. Each hyperedge records its own EMA advantage $w(e_{j})$, updated by Eq.~(4) when its members are again retrieved together for a high-reward question.

\noindent\textbf{Advantage-Weighted Reciprocal Rank Fusion.} A query is embedded by $f$ and matched against the node and hyperedge banks; a reciprocal-rank-fusion scorer~\citep{RRF} boosts each node by its advantage, so high-quality skills overtake more similar but low-utility ones in the ranking~\citep{RAG-original,RAGSurvey,LightRAG}:
\begin{equation}
\mathrm{score}(v_{i};q)\;=\;\frac{1\,+\,\tfrac{1}{2}(u_{i}\!+\!1)}{k_{\mathrm{rrf}}\,+\,\mathrm{rank}_{f}(v_{i};q)}\,,\qquad \mathcal{S}_{q}^{\mathrm{pool}}\;=\;\mathrm{Top}\text{-}k_{\cos}(\mathcal{V})\,\cup\,\mathrm{expand}_{q}(\mathcal{E})\,,
\end{equation}
where $\mathrm{rank}_{f}(v_{i};q)$ is the rank of $v_{i}$ under cosine similarity $\langle f(q),f(v_{i})\rangle$, $k_{\mathrm{rrf}}\!=\!60$, $k\!=\!100$ nearest nodes form the candidate pool, and $\mathrm{expand}_{q}(\mathcal{E})$ adds the members of hyperedges matching $q$, so that in libraries larger than the pool, skills composed with a relevant skill still enter it. Flat dictionaries~\citep{TrainingFreeGRPO,SkillCraft,AutoSkill} and frontier filters~\citep{EvoSkill} lack this per-item quality prior and joint composition.

\textbf{Proposition 2.} \textit{Advantage-weighted ranking places high-utility skills above merely similar ones.}
\vspace{-2mm}
\begin{proof}
We provide experimental results in \S\ref{sec:rq2} and \S\ref{sec:rq4} and proofs in Appendix~\ref{app:proof2}.
\end{proof}

\noindent\textbf{Similarity Floor, Capacity Cap, and Zero-Skill Fallback.} Injecting marginal skills from a small, noisy library can be worse than injecting none. A floor $\tau$ discards candidates whose cosine to $q$ is below it, a cap $N_{\max}$ bounds their number, and an empty result falls back to the skill-free prompt:
\begin{equation}
\mathcal{S}_{q}\;=\;\mathrm{Top}\text{-}N_{\max}\!\Bigl(\,\mathrm{score}(\,\cdot\,;q)\,,\;\;\bigl\{\,v_{i}\!\in\!\mathcal{S}_{q}^{\mathrm{pool}}\,:\,\langle f(q),f(v_{i})\rangle\!\geq\!\tau\,\bigr\}\,\Bigr)\,,\qquad N_{\max}\!=\!3\,,
\end{equation}
where $|\mathcal{S}_{q}|\!=\!0$ reduces $\pi_{\theta}(\cdot\!\mid\!q,m,\mathcal{S}_{q})$ to $\pi_{\theta}(\cdot\!\mid\!q,m)$; $\tau$ is set per dataset (App.~\ref{app:perdataset}). Retrieved skills enter as a strategy hint (App.~\ref{app:prompts}) or a terse list. The protocol skill of C1, when enabled, is added for every question without gating and re-injected after each tool turn with the remaining budget.

\textbf{Proposition 3.} \textit{Gating with zero-skill fallback never harms the skill-free pipeline in expectation.}
\vspace{-2mm}
\begin{proof}
We provide experimental results in \S\ref{sec:rq2} and \S\ref{sec:rq5} and proofs in Appendix~\ref{app:proof3}.
\end{proof}

\subsection{C3: Execution Protocol}
\label{sec:exec}

\noindent\textbf{Tool Calls and Final Answer.} C3 decides \emph{how} the frozen backbone executes a rollout; it never edits $\mathcal{H}$, and its options are set per dataset (App.~\ref{app:perdataset}). A \emph{concrete tool prompt} names the real tool tag (e.g.\ \texttt{<text\_search>}) instead of a \texttt{TOOL\_TAG} placeholder, and a \emph{lenient parser} also executes copied placeholder forms such as \texttt{<TOOL\_TAG>text\_search>q}. If a rollout ends without a closed \texttt{<answer>} tag (e.g.\ once the tool budget is spent), a \emph{final-answer turn} asks once more. An \emph{answer suffix} requests the final integer without \texttt{\textbackslash boxed\{\}} on math and the option letter on multiple choice.

\noindent\textbf{Evidence and Voting.} \emph{Question pre-search} sends the question verbatim to the \texttt{text\_search} tool once, outside the tool budget; \emph{locate pages} adds the text of the pages the question names and of the top BM25~\citep{BM25} pages of the question's own PDF, with no API call; on Omni, \emph{time grounding} lists the keyframe times and, on multiple choice, an \emph{option check} asks the model to weigh every option against frames and audio. \emph{Self-consistency}~\citep{SelfConsistency} adds $5$ samples at temperature $0.7$ and keeps their most frequent answer, or the greedy one on a tie or when no sample answers. C3 also sets per-dataset token budgets, and on five datasets its options also apply to the training rollouts that build the library, so the library is learned under the same protocol that serves its test questions (App.~\ref{app:perdataset}).

\noindent\textbf{Closing the Loop.} A training iteration on $\mathcal{B}_{t}$ retrieves $\mathcal{S}_{q}\!\leftarrow\!\mathcal{R}(q,m,\mathcal{H}_{t})$, generates $n$ rollouts under frozen $\theta$, records their failure types, computes advantages $\{\hat{A}_{q}\}$, admits the probe survivors of $\Phi\!\circ\!\Psi$, and updates $\mathcal{H}_{t}\!\to\!\mathcal{H}_{t+1}$ by EMA, hyperedge growth and pruning, so an auditable delta $\Delta\mathcal{H}$ replaces the GRPO gradient step $\Delta\theta$. At evaluation the library is frozen, and C2 and C3 serve each test question. Defaults are $(n,K,\eta,k_{\mathrm{rrf}},N_{\max},\rho,u_{\min})\!=\!(2,2,0.1,60,3,0.5,-0.5)$.

\clearpage

\begin{table}[t]
\centering
\caption{Main results on the 11 datasets of HyperSkill-Bench (\S\ref{sec:exp-setup}; token-level F1\,\%, CJK-aware scorer, full-set mean of one run). Subscripts: change vs.\ Zero-shot on the same dataset (\textcolor{green!55!black}{green $+$}, \textcolor{red!75!black}{red $-$}, \textcolor{gray}{grey $\pm0.0$}); \textbf{bold}: best, \underline{underline}: second best per row. All columns share the frozen \texttt{gemini-2.5-flash-lite} backbone, the same tools and scoring rule; HyperSkill is ours.}
\label{tab:main}
\setlength{\tabcolsep}{3pt}\renewcommand{\arraystretch}{1.15} 
\newcommand{\dsub}[2]{\textcolor{#1}{\raisebox{-1.6pt}{\fontsize{8}{8}\selectfont$#2$}}}%
\resizebox{\linewidth}{!}{%
\begin{tabular}{cl|cccccccc}
\toprule
\textbf{Modality} & \textbf{Dataset} & \textbf{Zero-shot} & \textbf{T.F.GRPO} & \textbf{SkillCraft} & \textbf{AutoSkill} & \textbf{EvoSkill} & \textbf{M.-Skills} & \textbf{XSkill} & \textbf{HyperSkill} \\
\midrule
\rowcolor{cyan!6}                                    & AIME2024     & 10.0 & 10.0\dsub{gray}{\pm0.0} & \underline{13.3}\dsub{green!55!black}{+3.3} & \underline{13.3}\dsub{green!55!black}{+3.3} & 10.0\dsub{gray}{\pm0.0} & \underline{13.3}\dsub{green!55!black}{+3.3} & 6.7\dsub{red!75!black}{-3.3} & \textbf{40.0}\dsub{green!55!black}{+30.0} \\
\rowcolor{cyan!6}                                    & AIME2025     & \underline{33.3} & 26.7\dsub{red!75!black}{-6.6} & 30.0\dsub{red!75!black}{-3.3} & \underline{33.3}\dsub{gray}{\pm0.0} & \underline{33.3}\dsub{gray}{\pm0.0} & 26.7\dsub{red!75!black}{-6.6} & 30.0\dsub{red!75!black}{-3.3} & \textbf{53.3}\dsub{green!55!black}{+20.0} \\
\rowcolor{cyan!6} \multirow{-3}{*}{\textbf{Text}}    & WebWalkerQA  & \underline{20.7} & 16.9\dsub{red!75!black}{-3.8} & 16.5\dsub{red!75!black}{-4.2} & \underline{20.7}\dsub{gray}{\pm0.0} & 18.6\dsub{red!75!black}{-2.1} & 16.3\dsub{red!75!black}{-4.4} & 14.8\dsub{red!75!black}{-5.9} & \textbf{23.3}\dsub{green!55!black}{+2.6} \\
\midrule
\rowcolor{blue!7}                                    & FVQA         & 21.1 & 27.0\dsub{green!55!black}{+5.9} & 17.8\dsub{red!75!black}{-3.3} & \underline{31.3}\dsub{green!55!black}{+10.2} & 23.9\dsub{green!55!black}{+2.8} & 19.5\dsub{red!75!black}{-1.6} & 26.6\dsub{green!55!black}{+5.5} & \textbf{41.7}\dsub{green!55!black}{+20.6} \\
\rowcolor{blue!7}                                    & InfoSeek     & 22.8 & 27.4\dsub{green!55!black}{+4.6} & 15.2\dsub{red!75!black}{-7.6} & 22.3\dsub{red!75!black}{-0.5} & 16.8\dsub{red!75!black}{-6.0} & 22.7\dsub{red!75!black}{-0.1} & \underline{33.4}\dsub{green!55!black}{+10.6} & \textbf{44.8}\dsub{green!55!black}{+22.0} \\
\rowcolor{blue!7}                                    & SimpleVQA    & 26.5 & 19.3\dsub{red!75!black}{-7.2} & 19.8\dsub{red!75!black}{-6.7} & 27.7\dsub{green!55!black}{+1.2} & 33.7\dsub{green!55!black}{+7.2} & 28.6\dsub{green!55!black}{+2.1} & \underline{33.8}\dsub{green!55!black}{+7.3} & \textbf{37.4}\dsub{green!55!black}{+10.9} \\
\rowcolor{blue!7} \multirow{-4}{*}{\textbf{VL-Web}}  & LiveVQA      & 55.0 & 56.3\dsub{green!55!black}{+1.3} & 59.9\dsub{green!55!black}{+4.9} & \underline{64.4}\dsub{green!55!black}{+9.4} & 59.4\dsub{green!55!black}{+4.4} & 58.5\dsub{green!55!black}{+3.5} & 59.6\dsub{green!55!black}{+4.6} & \textbf{65.7}\dsub{green!55!black}{+10.7} \\
\midrule
\rowcolor{teal!10}                                    & DocBench     & 32.8 & 26.1\dsub{red!75!black}{-6.7} & 16.0\dsub{red!75!black}{-16.8} & 32.9\dsub{green!55!black}{+0.1} & 32.8\dsub{gray}{\pm0.0} & \underline{35.1}\dsub{green!55!black}{+2.3} & 33.1\dsub{green!55!black}{+0.3} & \textbf{39.2}\dsub{green!55!black}{+6.4} \\
\rowcolor{teal!10} \multirow{-2}{*}{\textbf{VL-Doc}}  & MMLongB.     & 19.4 & 22.6\dsub{green!55!black}{+3.2} & 9.8\dsub{red!75!black}{-9.6} & 12.4\dsub{red!75!black}{-7.0} & 15.4\dsub{red!75!black}{-4.0} & 10.1\dsub{red!75!black}{-9.3} & \underline{23.0}\dsub{green!55!black}{+3.6} & \textbf{32.1}\dsub{green!55!black}{+12.7} \\
\midrule
\rowcolor{orange!9}                                    & OmniVideoB.  & 16.0 & 6.4\dsub{red!75!black}{-9.6} & 14.9\dsub{red!75!black}{-1.1} & 12.2\dsub{red!75!black}{-3.8} & 16.0\dsub{gray}{\pm0.0} & 5.1\dsub{red!75!black}{-10.9} & \underline{17.2}\dsub{green!55!black}{+1.2} & \textbf{31.5}\dsub{green!55!black}{+15.5} \\
\rowcolor{orange!9} \multirow{-2}{*}{\textbf{Omni}}    & WorldSense   & 43.8 & 47.5\dsub{green!55!black}{+3.7} & 43.8\dsub{gray}{\pm0.0} & 43.8\dsub{gray}{\pm0.0} & 43.8\dsub{gray}{\pm0.0} & \underline{49.1}\dsub{green!55!black}{+5.3} & 40.6\dsub{red!75!black}{-3.2} & \textbf{53.1}\dsub{green!55!black}{+9.3} \\
\midrule
\rowcolor{red!12} \multicolumn{2}{c|}{\textbf{Avg.}} & 27.4 & 26.0\dsub{red!75!black}{-1.4} & 23.4\dsub{red!75!black}{-4.0} & 28.6\dsub{green!55!black}{+1.2} & 27.6\dsub{green!55!black}{+0.2} & 25.9\dsub{red!75!black}{-1.5} & \underline{29.0}\dsub{green!55!black}{+1.6} & \textbf{42.0}\dsub{green!55!black}{+14.6} \\
\bottomrule
\end{tabular}%
}
\end{table}

\section{Experiments}
\label{sec:exp}

We answer six RQs: \textbf{RQ1:} Does HyperSkill rank first on every dataset? \textbf{RQ2:} What does each component add? \textbf{RQ3:} Does counterfactual distillation keep a compact library of useful skills? \textbf{RQ4:} How does the hypergraph evolve under retrieval feedback? \textbf{RQ5:} How sensitive is gating to $\tau$, $N_{\max}$ and library size? \textbf{RQ6:} What are HyperSkill's training cost, library size and latency?

\subsection{Experimental Setup}\label{sec:exp-setup}

\begin{wrapfigure}[13]{r}{0.42\textwidth}
\vspace{-4mm}
\centering
\includegraphics[width=0.40\textwidth]{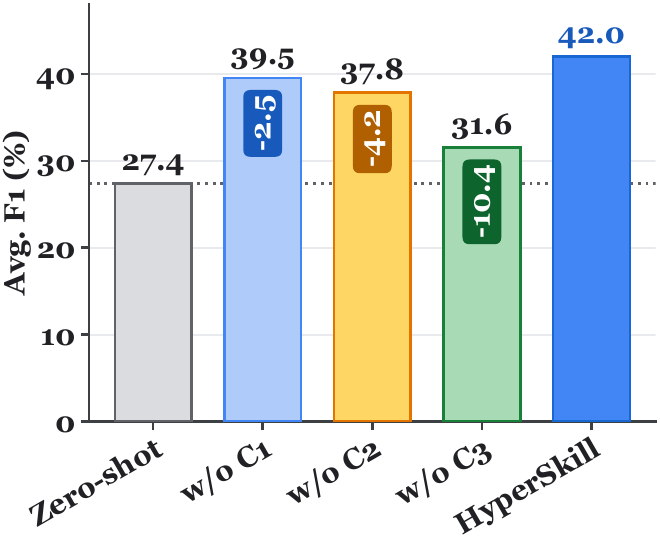}%
\vspace{-2mm}
\caption{Average F1 over the 11 datasets; white numbers: change vs.\ HyperSkill.}\label{fig:rq2}
\vspace{-3mm}
\end{wrapfigure}

\textbf{Datasets and Evaluation Metrics.} We evaluate on \emph{HyperSkill-Bench}, a collection of $11$ public datasets in four regimes (Appendix~\ref{app:dataset}): Text reasoning (AIME2024/2025, WebWalkerQA), VL-Web (FVQA, InfoSeek, SimpleVQA, LiveVQA), VL-Doc (DocBench, MMLongBench-Doc) and Omni (OmniVideoBench, WorldSense). The metric is token-level $F_1$~\citep{SQuAD} with answer-shape awareness and a CJK-aware tokeniser that splits Chinese into characters; every cell of Table~\ref{tab:main} and Figure~\ref{fig:rq2} is the full-set mean over the dataset's test questions ($30$ for AIME2024/2025, $64$ otherwise; $636$ in total), from one evaluation run per cell, scored by the same rule for every column (Appendix~\ref{app:eval}).

\textbf{Baselines and Ablations.} We compare against bare \emph{Zero-shot}~\citep{LLMsurvey} (same tools, no library or C3) and the $6$ training-free skill methods of Table~\ref{tab:comparison}: TrainingFreeGRPO~\citep{TrainingFreeGRPO} (experience pool), EvoSkill~\citep{EvoSkill} (frontier set), SkillCraft~\citep{SkillCraft} and AutoSkill~\citep{AutoSkill} (code, tip cards), Memento-Skills~\citep{Memento-Skills} and XSkill~\citep{XSkill} (memory). Each baseline builds its library on the same $128$ training questions by re-implementing its published procedure (Appendix~\ref{app:baseline}), likewise without C3. Three ablations reuse the trained library (Appendix~\ref{app:b-abl}): \emph{w/o C1} replaces all distilled content by a generic protocol line, \emph{w/o C2} also drops that line, and \emph{w/o C3}, applied on its own, only resets the C3 execution options to their defaults.

\textbf{Implementation Details.} Every method shares the frozen \texttt{gemini-2.5-flash-lite}~\citep{Gemini2.5} backbone with native vision/audio interfaces and the \texttt{gemini-embedding-2-preview}~\citep{GeminiEmbedding} encoder $f$, both via OpenRouter. Each HyperSkill library is trained for $1$ epoch on $128$ training questions per dataset (deduplicated, seed $42$) with batch size $4$, $n\!=\!2$ rollouts per question, seed $1$ and the defaults of \S\ref{sec:method}. Three library-run settings (R1--R3) differ in rollout temperature, early stopping, hint synthesis, probe sampling and training-time floor. Each dataset's run, floor $\tau$ and C3 options are selected on held-out training questions, and each library is then retrained on all $128$ questions (Appendix~\ref{app:perdataset}).

\textbf{Tool Stack and Decoding.} Every method uses the same regime tools: a \texttt{text\_search} tool for WebWalkerQA and VL-Web, served by \texttt{gemini-3.8-flash}; images passed to the backbone; a BM25 index over the question's own PDF for VL-Doc; and uniformly sampled keyframes for Omni, $8$ on OmniVideoBench and $32$ plus the audio track on WorldSense. Each question gets at most $3$ (Text), $5$ (VL) or $8$ (Omni) rounds and $2$ tool calls. Decoding is greedy at temperature $0$ with seed $1$. The baselines use $1024$ new tokens ($2048$ on AIME2025, $4096$ on WorldSense) and retrieve without a similarity floor; HyperSkill's token budgets and votes belong to C3. Variance: Appendix~\ref{app:eval}.

\clearpage


\subsection{Main Results (RQ1)}\label{sec:rq1}

\textbf{Overall.} HyperSkill is best or tied on all $11$ datasets (Table~\ref{tab:main}) and lifts the average from $29.0$ for the strongest baseline, XSkill, to $42.0$: a $+13.0$ absolute / $+44.9\%$ relative gain, and $+14.6$ over Zero-shot ($27.4$). The six baselines lie within $23.4$--$29.0$, $-4.0$ to $+1.6$ around Zero-shot, and each falls below Zero-shot on at least three datasets. HyperSkill is above Zero-shot on every row.

\textbf{Per-dataset.} The margin over the best other column is largest on AIME2024 ($+26.7$ over SkillCraft, AutoSkill and Memento-Skills), AIME2025 ($+20.0$ over three columns) and OmniVideoBench ($+14.3$ over XSkill), and is $+11.4$ on InfoSeek, $+10.4$ on FVQA and $+9.1$ on MMLongBench-Doc. The other five margins are $+1.3$ to $+4.1$, and their paired bootstrap 95\% intervals include zero (Appendix~\ref{app:eval}), so we conservatively count these five as ties. RQ2 then ablates C1, C1+C2, or C3 alone to measure how much each of the three components adds to the 11-dataset average.

\textbf{Per-regime.} By regime, HyperSkill's mean per-dataset lead over the best other column is $+16.4$ on Text, $+9.2$ on Omni, $+6.7$ on VL-Web and $+6.6$ on VL-Doc, so HyperSkill leads by more than $6$ F1 points in every regime, with the largest lead on the Text regime.

\subsection{Ablation Study (RQ2)}\label{sec:rq2}

The three ablations (Figure~\ref{fig:rq2}) lower the 11-dataset average from $42.0$ to $39.5$ (w/o C1), $37.8$ (w/o C2) and $31.6$ (w/o C3), and every variant stays well above Zero-shot ($27.4$); since every one of the ablations lowers the average, each of the three components is effective. The components act at different stages: C1 distils verified skills and a protocol skill from training failures, C2 delivers them to each test question behind the gate, and C3 makes the frozen backbone call its tools and commit to an answer. Only the full HyperSkill, which combines all three components, reaches the best 11-dataset average, and every ablated variant still exceeds all six baselines ($23.4$--$29.0$), so HyperSkill's lead over the baselines does not depend on any single one of its components.

\subsection{Counterfactual Distillation Effectiveness (RQ3)}\label{sec:rq3}

\begin{figure}[h]
\centering
\vspace{-2mm}
\includegraphics[height=3.15cm]{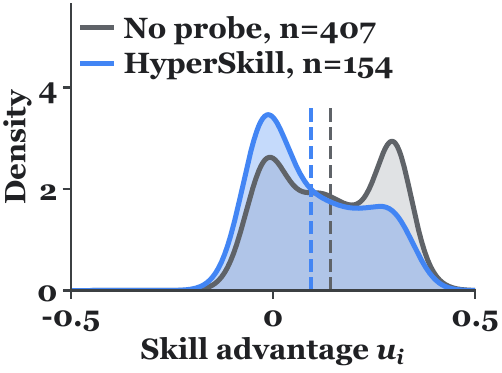}\hfill
\includegraphics[height=3.15cm]{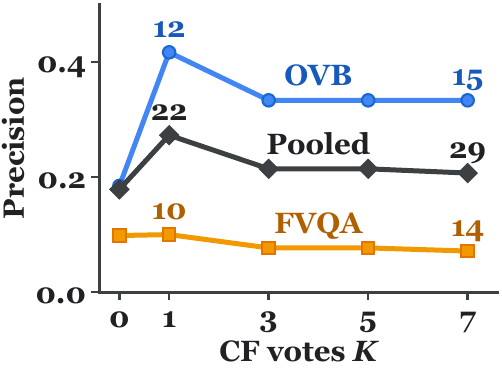}\hfill
\includegraphics[height=3.15cm]{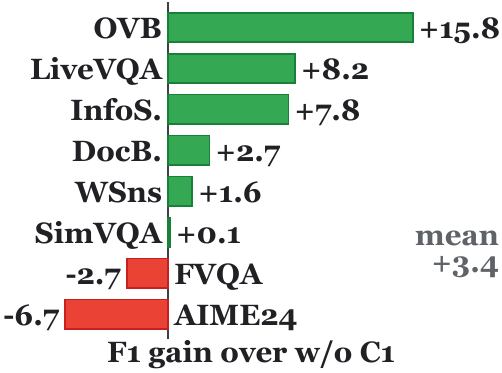}%
\vspace{-2mm}
\caption{(a) $u_{i}$ KDEs of HyperSkill vs.\ libraries trained without the probe; (b) probe precision vs.\ $K$ (labels: \#accepted; AIME24 none); (c) F1 gain over w/o C1 on the $8$ differing datasets.}\label{fig:rq3}
\end{figure}

Figure~\ref{fig:rq3} examines the counterfactual step-level distillation of \S\ref{sec:distill}, asking whether it keeps a compact library, whether the probe admits useful skills, and whether the distilled skills help at test time.

\textbf{Distilled skills are few and useful.} Pooled over the 11 libraries, HyperSkill admits $154$ skills, $38\%$ of the $407$ skills of the libraries trained without the probe (panel a), so the probe keeps the library compact. They also help: against w/o C1 (panel c), which swaps in a generic protocol line, HyperSkill gains on $6$ of the $8$ datasets where the two configurations differ (mean $+3.4$ F1), led by OmniVideoBench, LiveVQA and InfoSeek; the other three share one configuration.

\textbf{A single verification vote sharpens skill admission.} On $96$ of the $128$ training questions each of AIME2024, FVQA and OmniVideoBench (panel b), we take the $140$ (question, skill) candidates where the no-skill run fails (F1$\,<\!0.5$) and a skill is retrieved, and count a candidate as useful when injecting that skill flips the question to a success. Without the probe ($K\!=\!0$) $25$ of $140$ candidates are useful ($0.179$). One counterfactual vote raises pooled precision to $0.273$ ($6/22$) and, on OmniVideoBench, from $0.185$ to $0.417$ ($5/12$). Votes at temperature $0.1$ are nearly deterministic: a single vote already captures the filtering gain and every $K\!\geq\!1$ keeps pooled precision above the no-probe rate; hence the libraries use one vote (R2/R3) or two votes on one failure (R1).

\subsection{Hypergraph Evolution under Retrieval Feedback (RQ4)}\label{sec:rq4}

\begin{figure}[h]
\centering
\vspace{-2mm}
\includegraphics[height=4.5cm]{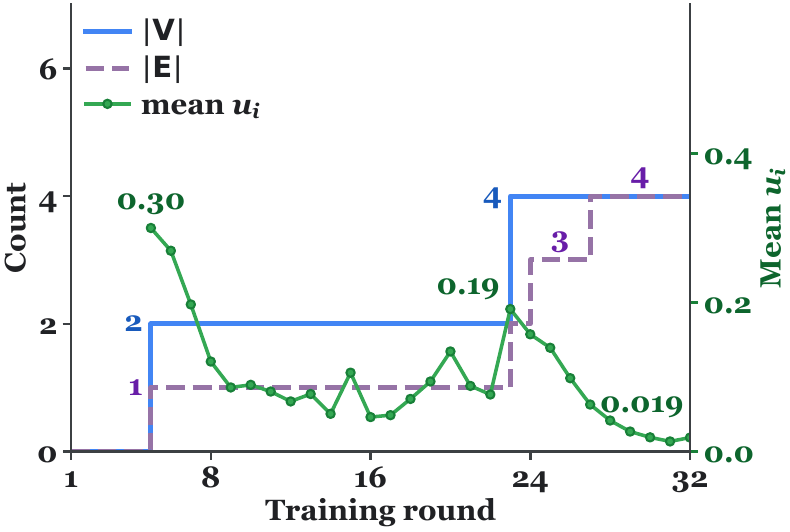}\hfill
\includegraphics[height=4.5cm]{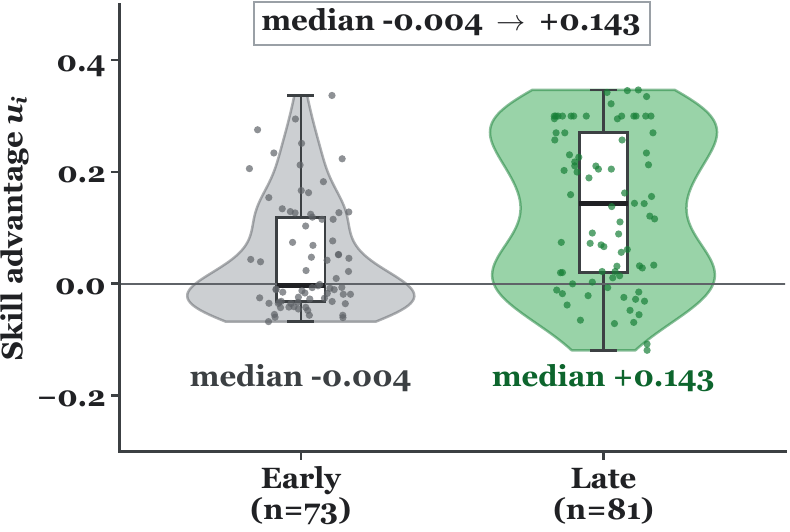}%
\vspace{-2mm}
\caption{(a) Library size ($|\mathcal{V}|$, $|\mathcal{E}|$) and mean $u_{i}$ over the $32$ rounds of an AIME2024 training re-run; (b) violins of $u_{i}$ for the early- and late-admitted halves pooled over the 11 final libraries.}\label{fig:rq4}
\end{figure}

Figure~\ref{fig:rq4} traces the weighted hypergraph of \S\ref{sec:hypergraph} over training, asking how retrieval feedback grows the library and its connectivity and how the advantages $u_{i}$ of the admitted skills evolve.

\textbf{Skills accumulate early and stabilise.} In an AIME2024 training re-run with per-round logging ($32$ rounds of $4$ questions; panel a), the library grows from $0$ to $2$ skills at round $5$ and to $4$ at round $23$, where it stays through round $32$; the hyperedges grow from $0$ to $1$ and reach $4$ at round $27$, when the library, taken as a skill--hyperedge graph, reaches $8$ nodes and $10$ links. The final $4$ skills match the size of the AIME2024 library used in Table~\ref{tab:main}. Admitted skills start at $u_{i}\!=\!0.3$ and the EMA of Eq.~(4) pulls them toward the batch-normalised advantage, so the mean $u_{i}$ moves from $0.30$ (round $5$) to $0.19$ (round $23$) and $0.019$ (round $32$) and stays positive throughout the training re-run.

\textbf{Advantages keep being refined after admission.} In the 11 libraries (panel b), the earlier-admitted half of each library, updated most often, has median $u_{i}\!\approx\!0$ ($73$ skills) and the later half $+0.143$ ($81$ skills): the gap ($\approx\!0.15$) grows with the EMA updates that pull each $u_{i}$ from $0.3$ toward $\hat{A}_{q}$.


\subsection{Sensitivity of Gated Retrieval (RQ5)}\label{sec:rq5}

\begin{figure}[h]
\centering
\vspace{-2mm}
\includegraphics[height=3.15cm]{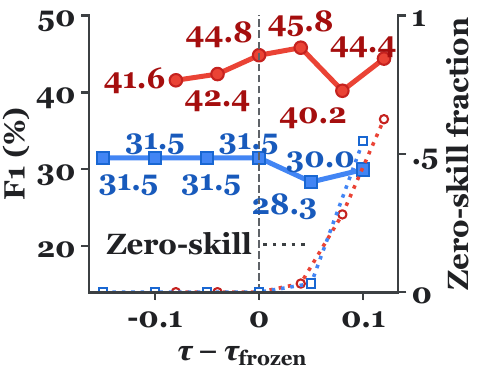}\hfill
\includegraphics[height=3.15cm]{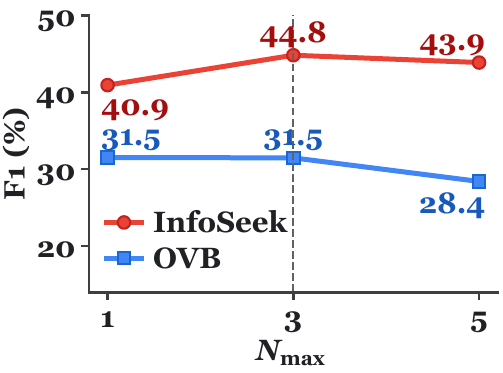}\hfill
\includegraphics[height=3.15cm]{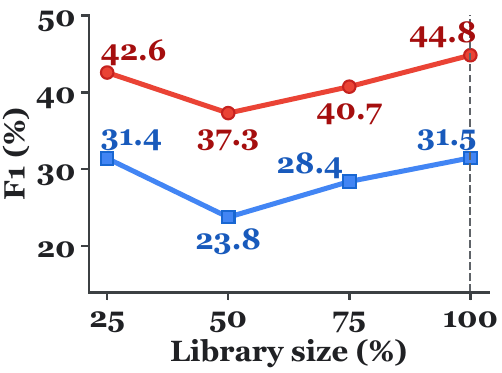}%
\vspace{-2mm}
\caption{(a) F1 and zero-skill fraction vs.\ the offset of $\tau$ from its frozen value; (b) F1 vs.\ $N_{\max}$; (c) F1 of gated retrieval on nested library subsets. Grey lines: frozen settings.}\label{fig:rq5}
\end{figure}

Figure~\ref{fig:rq5} varies one setting at a time around the frozen configuration of InfoSeek and OmniVideoBench, which cover the image-based VL-Web and the video-based Omni regimes ($64$ test questions per cell, single runs). The floor $\tau$ takes six values around its frozen value on each dataset (InfoSeek on the hub-corrected scale of Appendix~\ref{app:perdataset}), $N_{\max}\!\in\!\{1,3,5\}$, and the library subsets are nested random subsets with $25$, $50$, $75$ and $100\%$ of the skills, where a hyperedge is kept only if all its member skills are kept; ungated retrieval sets $\tau\!=\!-9$. This is a post-hoc sensitivity analysis on the test split; it does not change the frozen per-dataset configuration or the reported results of Table~\ref{tab:main}.

\textbf{The frozen settings lie in a stable region.} Around the frozen floor (panel a), InfoSeek stays between $40.20$ and $45.80$ F1 ($44.82$ at its hub-corrected $\tilde{\tau}\!=\!-0.02$), and OmniVideoBench gives $31.47$ at every $\tau\!\leq\!0.55$, where injected skills are identical, and $28.33$--$29.95$ at the two larger values.

\textbf{The floor controls injection.} The zero-skill fraction is $0$ at the frozen $\tau$ on both datasets and rises to $0.63$ and $0.55$ at the largest $\tau$, where on average $0.70$ and $0.78$ skills are injected per question.

\textbf{A small capacity cap suffices.} $N_{\max}\!=\!3$ is best or near-best on both datasets (panel b): $44.82$ vs $40.93$ at $1$ and $43.87$ at $5$ on InfoSeek, and $31.47$ vs $31.51$ at $1$ and $28.40$ at $5$ on OmniVideoBench, with $1$, $3$ and up to $5$ skills injected per question at the three cap values, respectively.

\textbf{Gated, the full library is best.} With gating (panel c), the full library gives the highest F1 on both datasets ($44.82$ on InfoSeek, $31.47$ on OmniVideoBench), $+4.1$ and $+3.1$ over the $75\%$ subset; at full size, gated and ungated retrieval give $44.82$ vs $44.64$ on InfoSeek and $31.47$ vs $31.47$ on OmniVideoBench, where the injected skills coincide, so gating lowers F1 on neither dataset.

\subsection{Cost Analysis (RQ6)}\label{sec:rq6}

\begin{figure}[h]
\centering
\vspace{-2mm}
\includegraphics[height=4.6cm]{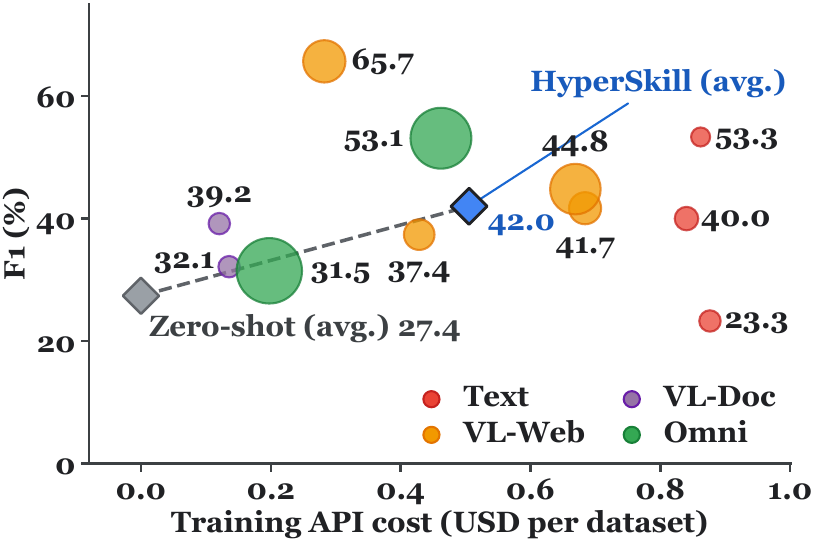}\hfill
\includegraphics[height=4.6cm]{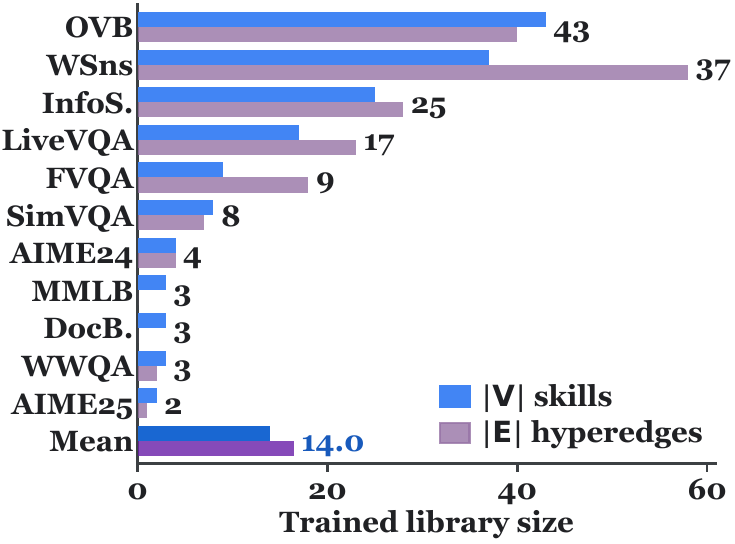}%
\vspace{-2mm}
\caption{(a) Training API cost vs.\ F1 per dataset, coloured by regime, bubble area $\propto|\mathcal{V}|$; (b) library size per dataset: skills $|\mathcal{V}|$ (dark) and hyperedges $|\mathcal{E}|$ (light), with the mean in the last row.}\label{fig:rq6}
\end{figure}

Figure~\ref{fig:rq6} reports two deployment-relevant costs of HyperSkill on all $11$ datasets: the training-API spend to build each library and the size of the learned library in skills and hyperedges; we also compare the per-question inference latency of HyperSkill with Zero-shot on three datasets.

\textbf{Cheap to train, compact to store.} HyperSkill's library costs $\$0.12$--$\$0.88$ per dataset to train (mean $\$0.51$, total $\$5.56$, of which $\$2.48$ is knowledge search) and holds $2$--$43$ skills (mean $14.0$) and $0$--$58$ hyperedges; one test pass over all 11 datasets costs $\$2.34$. Panel (a) adds the Zero-shot ($27.4$) and HyperSkill ($42.0$) averages, and panel (b) sorts the datasets by library size, with the mean in the last row; the two Omni libraries are the largest ($43$ and $37$ skills), and the $154$ skills pooled over all $11$ libraries are $38\%$ of the $407$ skills of the libraries trained without the probe (\S\ref{sec:rq3}). Training needs no gradient step: each library is built from API calls alone, and per question the counterfactual check of C1 adds at most $2|\widetilde{\mathcal{V}}_{q}|$ verification calls at $K\!=\!2$ (Eq.~(\ref{eq:a-calls})), which take the place of the backward pass of standard GRPO (Appendix~\ref{app:prompts}) and are already included in the training costs above.

\textbf{Inference latency stays close to Zero-shot.} Measured for HyperSkill and Zero-shot on $16$ test questions per timed dataset with the same harness, mean latency per question is $14.98$ vs $11.75$\,s on AIME2024, $12.78$ vs $14.56$\,s on FVQA and $3.56$ vs $3.66$\,s on DocBench: HyperSkill adds at most about $3$\,s (on AIME2024) and is on par or slightly faster elsewhere, since API, tool and network time dominate and C1 and C3 add only short instruction texts to the prompt (Appendix~\ref{app:prompts}). The timed datasets cover three regimes (Text, VL-Web and VL-Doc), each with its own tools (\S\ref{sec:exp-setup}).

\section{Conclusion}\label{sec:conclusion}

HyperSkill is a \emph{Training-Free Omnimodal GRPO} framework that evolves a \emph{Hypergraph-Indexed Skill-Library} through training-time distillation (C1), gated hypergraph retrieval (C2) and an execution protocol (C3). Without parameter updates, it is best or tied on all $11$ datasets (avg.\ $42.0$, $+13.0$ over the strongest baseline), and every skill stays readable and editable. Each component raises the average, counterfactual distillation keeps a compact library ($154$ vs $407$ skills), and the gate settings lie in a stable region. A library costs at most $\$0.88$ to train, a full test pass costs $\$2.34$, and latency stays close to Zero-shot. Future work includes pairing the library with lightweight parameter updates, sharing it across related datasets and domains, and extending C3 to new tool types and modalities.

\clearpage
\raggedbottom
\section*{Acknowledgments}
This work is supported by the National Research Foundation, Singapore, under its Frontier Competitive Research Programme (NRF-F-CRP-2024-0005). Any opinions, findings and conclusions or recommendations expressed in this material are those of the authors and do not reflect the views of National Research Foundation, Singapore. The authors declare no competing interests.

\begingroup
\setlength{\bibsep}{8pt minus 2pt}
\bibliography{custom}{}
\bibliographystyle{plain}
\endgroup


\newpage
\flushbottom
\appendix
\section*{Appendix} 
\section{Prompts Used in HyperSkill}\label{app:prompts}

This appendix lists the LLM prompts invoked by HyperSkill, mirroring the prompt-card presentation of HyperGraphRAG~\citep{HyperGraphRAG}. 
Prompts are grouped by the role they play in the loop of \S\ref{sec:method}: \textbf{(a) Frozen-Backbone Inference} (\S\ref{sec:method}) consumed by every skill-augmented rollout, \textbf{(b) C1 Skill Distillation} (\S\ref{sec:distill}) that admits counterfactually checked skill nodes, \textbf{(c) C2 Gated Retrieval Injection} (\S\ref{sec:hypergraph}) that surfaces a question-aware strategy hint, and \textbf{(d) the C1 protocol skill and the C3 execution texts} (\S\ref{sec:exec}). The weighted hypergraph evolution inside C2 is deliberately \emph{prompt-free}: per-node advantage EMA, hyperedge growth and the periodic $u_{\min}$ safeguard prune are closed-form operators on the rollout reward and never call $\pi_{\theta}$, so no prompt card appears for that block. Every prompt below is reproduced from the released code (markdown emphasis, dashes and symbols normalised to plain text), with \texttt{\{x\}} denoting a runtime placeholder; the user-message wrapper around each system prompt is kept compact since the system prompt already constrains the output schema.

\subsection{Frozen-Backbone Inference Prompts}

\noindent{\textbf{Prompt P1: Base Reasoning} (\texttt{REASONING\_PROMPT}, used on tool-free Text datasets (AIME)).} The answer-only base prompt of tool-free datasets; like P2/P3, it is used without retrieved skills whenever Eq.~(7) returns the empty skill set (the zero-skill fallback of \S\ref{sec:gating}), and with the P8 tip line otherwise.
\begin{shaded}
\footnotesize
\begin{verbatim}
Answer the following question. Output ONLY your final answer inside
<answer>...</answer> tags. Do NOT show your work.
\end{verbatim}
\end{shaded}

\noindent{\textbf{Prompt P2: Search Agent} (\texttt{SEARCH\_AGENT\_PROMPT}, used on Text/VL tool tasks with the \texttt{text\_search} / BM25 channels of \S\ref{sec:exp-setup}).} Drives the per-question multi-round loop in which the agent either commits to an answer or emits exactly one tool tag per round under the regime cap (3 rounds on Text, 5 on VL; Omni datasets use P3) and the shared budget of $2$ tool calls.
\begin{shaded}
\footnotesize
\begin{verbatim}
You are an expert research assistant equipped with search tools.
Analyze the question and decide whether to search for more information
or to provide a final answer directly.
{tool_desc}

You must respond in ONE of the following formats:

  Format A - Direct answer (when confident):
    <reason>...</reason>
    <answer>...</answer>

  Format B - Search for more information:
    <reason>...</reason>
    <TOOL_TAG>specific query</TOOL_TAG>

Replace TOOL_TAG with one of the available tools listed above.
\end{verbatim}
\end{shaded}
\noindent On WebWalkerQA, for every method, the last two lines of P2 read \texttt{<text\_search>specific query</text\_search>} and ``Put the search query itself between the \texttt{<text\_search>} and \texttt{</text\_search>} tags.''

\noindent{\textbf{Prompt P3: Omni Agent} (\texttt{OMNI\_AGENT\_PROMPT}, used on the Omni regime).} The omnimodal counterpart of P2: keeps the same two-format response schema while explicitly instructing the backbone to fuse text, image, audio transcript, and video keyframes in one multimodal call.
\begin{shaded}
\footnotesize
\begin{verbatim}
You are an expert multimodal reasoning assistant. You can process text,
images, audio transcripts, and video descriptions. Analyze all provided
modalities together to derive a precise answer.
{tool_desc}

You must respond in ONE of the following formats:

  Format A - Direct answer (when confident):
    <reason>...</reason>
    <answer>...</answer>

  Format B - Search for more information:
    <reason>...</reason>
    <TOOL_TAG>specific query</TOOL_TAG>

Replace TOOL_TAG with one of the available tools listed above.
\end{verbatim}
\end{shaded}

\noindent{\textbf{Prompt P4: After-Search Continuation} (\texttt{AFTER\_SEARCH\_PROMPT}).} Re-enters the agent into another reasoning round with the original question and the freshly retrieved evidence; iterated until the regime-specific round cap ($3$, $5$ or $8$) is reached or an \texttt{<answer>} tag closes the reasoning trace.
\begin{shaded}
\footnotesize
\begin{verbatim}
Based on the search results provided, continue your reasoning and either
search again (if needed) or provide the final answer.

Original question: {question}

{search_results}

Now provide your next step:
\end{verbatim}
\end{shaded}

\subsection{C1: Skill Distillation Prompts (\S\ref{sec:distill})}

\noindent{\textbf{Prompt P5: Contrastive Step Alignment} ($\Phi$ in Eq.~(2), \texttt{SKILL\_CONTRASTIVE\_EXTRACT}).} Operates on the success/failure split at threshold $\rho\!=\!0.5$: aligns numbered reasoning steps across the two buckets, marks divergence points, and harvests both atomic skills and multi-skill recipes that successes exhibit but failures jointly miss; recipes feed downstream hyperedge construction in \S\ref{sec:hypergraph}.
\begin{shaded}
\footnotesize
\begin{verbatim}
[system]
You are an expert at analyzing reasoning trajectories to extract reusable
skills. Your task: given a question with GOOD attempts (high reward) and
BAD attempts (low reward), perform step-level contrastive analysis to
find exactly WHERE and WHY the good attempts succeeded.
<Instructions>
1. Decompose each attempt into numbered reasoning steps.
2. Align steps across good and bad attempts - find corresponding steps
   that address the same sub-problem.
3. Identify divergence points - the specific steps where good and bad
   attempts DIFFER in strategy or correctness.
4. Extract atomic skills - from each divergence point, extract the
   generalizable skill that the good attempt used (NOT specific to
   this question).
5. Identify skill recipes - if multiple atomic skills worked TOGETHER
   in successful attempts but were missing together in failures, group
   them as a recipe.
</Instructions>
<Output Format>
<Step Alignment>
[Show the aligned steps of good vs bad attempts, marking divergence
 points with *]
</Step Alignment>
<Divergence Analysis>
[For each divergence point: what the good attempt did vs what the bad
 attempt did, and WHY the good approach worked]
</Divergence Analysis>
<Atomic Skills>
[One per line, extracted from divergence points:]
SKILL: <name (3-7 words)> | <what to do and when (1-2 sentences)>
</Atomic Skills>
<Skill Recipes>
[Combinations of atomic skills that must be used TOGETHER for success:]
RECIPE: <recipe name> | <skill_1 + skill_2 + ...> |
        <when to apply this combination>
</Skill Recipes>
</Output Format>

[user]
<Question>{question}</Question>
<Gold Answer>{gold_answer}</Gold Answer>
<Good Attempts>{good_attempts}</Good Attempts>
<Bad Attempts>{bad_attempts}</Bad Attempts>
Perform step-level contrastive analysis as instructed.
\end{verbatim}
\end{shaded}

\noindent{\textbf{Prompt P6: Counterfactual Verification} (\texttt{SKILL\_COUNTERFACTUAL\_VERIFY}; $K$-probe vote of Eq.~(3)).} Each candidate emitted by P5 is paired with up to $K\!=\!2$ failure probes of its own question (Appendix~\ref{app:perdataset}) and admitted only when a strict majority of probes return YES; the $\mathrm{md5}$ name identifier then deduplicates the increment $\mathcal{V}_{t}^{\mathrm{new}}$ against $\mathcal{V}_{t}$, so a repeated skill name is skipped.
\begin{shaded}
\footnotesize
\begin{verbatim}
[system]
You are evaluating whether a proposed reasoning skill would help an
agent that FAILED a question.
Given:
- A question the agent got WRONG
- The agent's FAILED reasoning trajectory
- A proposed SKILL
Judge: If the agent had known and applied this skill, would it likely
have answered correctly?
Output EXACTLY one line:
VERDICT: YES | <confidence 0-1> | <brief reason>
or
VERDICT: NO | <confidence 0-1> | <brief reason>

[user]
<Question>{question}</Question>
<Gold Answer>{gold_answer}</Gold Answer>
<Failed Trajectory>{failed_trajectory}</Failed Trajectory>
<Proposed Skill>{skill}</Proposed Skill>
Would this skill have helped? Output VERDICT line only.
\end{verbatim}
\end{shaded}

\subsection{C2: Gated Retrieval Injection Prompts (\S\ref{sec:hypergraph})}

\noindent{\textbf{Prompt P7: Strategy Hint Synthesis} (\texttt{SKILL\_SUMMARY\_PROMPT}).} After the advantage-weighted RRF of Eq.~(6) and the similarity/capacity gates of Eq.~(7) yield a non-empty $\mathcal{S}_{q}$, this prompt distils the retrieved skills into a question-aware $2$--$3$ sentence hint that emphasises \emph{how to combine} them rather than enumerating them, so that the context injected into the prompt stays short.
\begin{shaded}
\footnotesize
\begin{verbatim}
Given a question and a set of retrieved reasoning skills, write a concise
strategy hint (2-3 sentences) that synthesizes these skills specifically
for this question. Focus on HOW to combine them, not just listing them.
Question: {question}
Retrieved Skills:
{skills_raw}
Strategy hint (2-3 sentences, actionable):
\end{verbatim}
\end{shaded}
\noindent With the \emph{schema-constrained hint} (SimpleVQA), the following constraint is inserted before the last line of P7; the bracketed clause is added only in tool regimes, with \texttt{\{limit\}} the tool budget.
\begin{shaded}
\footnotesize
\begin{verbatim}
Constraints for the hint: it must not change the required output format
(the agent ends with a short final answer inside <answer>...</answer>, as
short as the question allows, never a full sentence when a name, number
or phrase suffices). [It must not ask for more than {limit} tool calls in
total, and the agent must always finish with a final <answer> even if a
search did not find everything.]
\end{verbatim}
\end{shaded}

\noindent{\textbf{Prompt P8: Skill-Augmented Wrapper} (\texttt{SKILL\_AUGMENTED\_PROMPT}).} The hint produced by P7 is prepended to the inference prompt of P1/P2/P3 as a single \emph{tip} line; whenever $|\mathcal{S}_{q}|\!=\!0$ no retrieved skill enters this wrapper, which then carries only the protocol skill or is skipped; this is the zero-skill fallback that Proposition~3 analyses, with no P7 call for such zero-skill questions.
\begin{shaded}
\footnotesize
\begin{verbatim}
Tip from prior experience: {skills_context}
\end{verbatim}
\end{shaded}

\subsection{C1 Protocol-Skill and C3 Execution Texts (\S\ref{sec:distill}, \S\ref{sec:exec})}\label{app:c3prompts}

\noindent These texts are reproduced verbatim from the released code (\texttt{strategies/protocol.py}, \texttt{prompting.py}, \texttt{presearch.py}, \texttt{locate\_pages.py}, \texttt{video.py}); \texttt{\{x\}} is a runtime placeholder and \texttt{[...]} names the condition under which a line is used; all remaining text is fixed.

\noindent{\textbf{Prompt P9: Protocol Skill} (C1, injected with the skills context).} Instruction lines for the failure types kept in training (at least $2$ occurrences; \texttt{no\_answer} adds none of its own), followed by one answer-form line with word limit \texttt{\{words\}} $=\max(200,\lfloor0.45\cdot\text{max new tokens}\rfloor)$.
\begin{shaded}
\footnotesize
\begin{verbatim}
Output protocol (learned from training failures): <lines below>
[malformed_tool] To call a tool, put the query inside that tool's own
  tag, exactly like <{tag}>your query</{tag}>; the tag name is the tool
  name (never write TOOL_TAG and never nest tags).
[over_budget or malformed_tool] Use at most {budget} tool calls in
  total; after the last one, or as soon as you can answer, stop calling
  tools and give your best final answer.
[tool_without_tools] No tool is available for this question: do not
  write any tool call, search request or TOOL_TAG; answer directly from
  the given input.
[math] Reason concisely (the whole response must stay well under
  {words} words) and end with the final integer inside
  <answer>...</answer>; do not use \boxed{}.
[short / default] Always end with your final answer inside
  <answer>...</answer>, as short as the question allows (a name, number,
  short phrase or option letter, not a full sentence).
[data form, option letters] Always end with only the option letter of
  your choice inside <answer>...</answer> (e.g. <answer>B</answer>).
[data form, long golds] Always end with your final answer inside
  <answer>...</answer>, written as one complete sentence (about {n}
  words) that restates the key terms of the question together with the
  answer.
[data form, frequent abstention] If the input does not contain the
  answer, answer exactly <answer>{frequent}</answer>.
\end{verbatim}
\end{shaded}

\noindent{\textbf{Prompt P10: Protocol Re-injection} (C1/C2, after each tool turn, inside the ``\texttt{-{}-{}- Updated Skills -{}-{}-}'' block).} Appended to every tool result while the protocol skill is on (line~7 of Algorithm~\ref{alg:hs-c3}), so that both the remaining tool budget and the required final-answer tag stay in view.
\begin{shaded}
\footnotesize
\begin{verbatim}
[calls left] Protocol: you may make {left} more tool call(s), written as
  <{tag}>your query</{tag}>; if the results above suffice, give the
  final answer inside <answer>...</answer> now.
[budget spent] Protocol: you have used all {budget} tool calls. Do not
  call a tool again; give your best final answer now inside
  <answer>...</answer>.
\end{verbatim}
\end{shaded}

\noindent{\textbf{Prompt P11: Final-Answer Turn} (C3).} \texttt{\{form\}} is `` (only the final integer)'' on math, `` (only the option letter)'' on multiple choice, and empty otherwise; the turn is sent at most once per rollout.
\begin{shaded}
\footnotesize
\begin{verbatim}
[budget spent or repeated query] Tool budget exhausted: no further tool
  call will be executed. Based on the information above, give your final
  answer now inside <answer>...</answer>{form}.
[otherwise] You have not given your final answer yet. Give it now inside
  <answer>...</answer>{form}.
\end{verbatim}
\end{shaded}

\noindent{\textbf{Prompt P12: Answer Suffix and Concrete Tool Prompt} (C3).} The suffix follows the question. The concrete tool prompt replaces the \texttt{TOOL\_TAG} lines of P2/P3 by the real tag and one example; without tools (Omni), it drops the search format instead; the lenient parser accepts placeholders.
\begin{shaded}
\footnotesize
\begin{verbatim}
[math] Reason step by step, then end your response with the final
  integer inside <answer>...</answer> (for example <answer>N</answer>
  for the integer N); do not use \boxed{}.
[multiple choice] End your response with only the letter of the correct
  option inside <answer>...</answer> (for example <answer>X</answer> for
  option X).
[concrete tool prompt, Format B] <{tag}>specific query</{tag}>
  The tag is the tool's own name. Example of a correct tool call:
  <text_search>Eiffel Tower height in metres</text_search>
\end{verbatim}
\end{shaded}

\noindent{\textbf{Prompt P13: Evidence Texts} (C3: question pre-search, locate pages, time grounding, option check).} Each text is added only when the dataset enables the matching C3 option (QP, PG$n$, TG, OC in Appendix~\ref{app:perdataset}); the pre-search call is not counted against the per-question tool budget.
\begin{shaded}
\footnotesize
\begin{verbatim}
[pre-search] Results of a knowledge search for the question itself (run
  automatically before your first step):
  [text_search Results]
  {results}
  Use these results if they answer the question; otherwise search
  further.
[locate pages] Candidate pages of {pdf}, located by document search for
  this question (text extracted from the PDF): [page k] {text} ...
  Answer from these pages. If they do not contain the answer, search the
  document again with <doc_search>...</doc_search>. Answer "Not
  answerable" only if the information is still not found after
  searching. Put the final answer inside <answer>...</answer>.
[time grounding] Video timing: the video lasts {T}. The {n} images are
  keyframes in temporal order, taken at image 1 = {t1}, ... The audio
  track covers the whole video on the same timeline (0:00 to {T}). For
  questions about when something happens, compare its time with the
  total length: the beginning is the first third, the middle the second
  third and the end the last third. Between two keyframes, rely on the
  audio and the neighbouring frames.
[option check] Before answering, check every option against what the
  frames show and what the audio says: for each option write one short
  line 'X: supported / contradicted / no evidence' with the frame time
  or the audio it relies on, then choose the best-supported option.
  Choose an option such as 'not sure', 'unknown', 'it is unclear' or
  'none of the above' only if every other option is contradicted. Never
  decline to answer: always end with exactly one option letter inside
  <answer>...</answer>, for example <answer>B</answer>.
\end{verbatim}
\end{shaded}

\subsection{Per-Baseline Injection Formats and Per-Question Call Budget}

\noindent{\textbf{Injection formats.}} Each training-free baseline of \S\ref{sec:exp-setup} builds its \emph{own} library (Appendix~\ref{app:baseline}) and injects its retrieved entries in its own surface form, prepended to the shared inference prompt. The forms, reproduced from the released baseline code with their retrieval depth in parentheses, are:
\begin{shaded}
\footnotesize
\begin{verbatim}
trainingfreegrpo : When solving problems, you MUST first carefully read
                   and understand the helpful instructions and
                   experiences: [<id>]. <experience> ...        (top-10)
skillcraft       : Reusable skills (procedures composed from your
                   operations; apply the most relevant one, then answer
                   in the required format): ```python <code>```  (top-3)
autoskill        : Relevant skill cards from prior experience:
                   * Skill: <name> -- <description>
                     * Tip: <tip> ...                             (top-3)
evoskill         : > [frontier-skill, val_gain=<gain>] <name>: <desc>
                      1. <instruction> ...                        (top-3)
mementoskills    : [Memory 1] Recall a similar past case:
                   <markdown skill file>                          (top-1)
xskill           : <> Cross-task experiences: - <experience> ...  (top-3)
                   <> Skill (<name>): 1. <step> ...               (top-1)
\end{verbatim}
\end{shaded}

\noindent{\textbf{Per-question LLM call budget.}} For one round-$t$ batch question $q$ with $n\!=\!2$ skill-augmented rollouts under regime cap $R\!\in\!\{3,5,8\}$, the prompts P1--P13 above together fire at most
\begin{equation}
\underbrace{\mathbf{1}\bigl[|\mathcal{S}_{q}|>0\bigr]+n\,(R{+}1)}_{\text{inference}}\;+\;\underbrace{1}_{\text{distillation}}\;+\;\underbrace{K\,|\widetilde{\mathcal{V}}_{q}|}_{\text{verification}}
\label{eq:a-calls}
\end{equation}
LLM calls: P7 once for hint synthesis when the gates of Eq.~(7) admit at least one skill, then P1/P2/P3 followed by up to $R{-}1$ P4 continuations and at most one final-answer turn per rollout; P5 on the $n$-rollout pool whenever at least one success and one failure coexist; and P6 once per (candidate, failure-probe) pair, i.e.\ at most $2|\widetilde{\mathcal{V}}_{q}|$ calls at $K\!=\!2$. P8 is a one-line wrapper that fires zero LLM calls. Compared to the backward pass of standard GRPO, this replaces $\Delta\theta$ with $\mathcal{O}(|\widetilde{\mathcal{V}}_{q}|)$ frozen-backbone calls per question, which benefit from prompt caching since the P5/P6 system messages are identical across calls and only the question, trajectories and candidate skill change per call.

\textbf{Reproducibility.} The prompts above, with Eqs.~(2)--(3) and the closed-form operators of Eqs.~(4)--(7), specify HyperSkill's LLM surface on the unified \texttt{gemini-2.5-flash-lite} backbone (\S\ref{sec:exp-setup}). Evaluation decodes at temperature $0$ (self-consistency samples at $0.7$); training rollouts use temperature $0.4$ in library runs R2/R3 and $0.7$ (Text) / $1.0$ (VL, Omni) in run R1, distillation $0.3$, verification $0.1$, and hint synthesis $0$. The default hyperparameters are $(n,K,\eta,k_{\mathrm{rrf}},N_{\max},\rho,u_{\min})\!=\!(2,2,0.1,60,3,0.5,-0.5)$ with protocol threshold $2$; the per-dataset floor $\tau$, injection variants and C3 options are listed in Appendix~\ref{app:perdataset}, together with the library run of each dataset.

\clearpage

\section{Proofs of Propositions}\label{app:proof}

This appendix proves Propositions~1--3 stated in \S\ref{sec:method}, formalising the three claims that ground counterfactual step-level skill distillation (C1, \S\ref{sec:distill}), the advantage-weighted hypergraph retrieval of C2 (\S\ref{sec:hypergraph}), and its gate with zero-skill fallback (C2, \S\ref{sec:gating}). Notation follows \S\ref{sec:method}: $\pi_{\theta}$ is the frozen omnimodal backbone, $\mathcal{H}\!=\!(\mathcal{V},\mathcal{E},w)$ the weighted skill hypergraph with per-node advantage $u_{i}\!\in\![-1,1]$, $\mathcal{R}$ the retriever, $f$ the shared encoder, and $r\!\in\![0,1]$ the token-level $F_{1}$ reward against $a^{*}$.

\subsection{Proof of Proposition 1}\label{app:proof1}
\textbf{Proposition 1.} \textit{Counterfactual distillation favours failure-rescuing skills over co-occurring ones.}\vspace{-2mm}
\begin{proof}
Fix a question $q\!\in\!\mathcal{B}_{t}$ and let $\widetilde{\mathcal{V}}_{q}$ be the candidate set returned by the contrastive aligner of Eq.~(2). Each candidate $\widetilde{v}$ has an unobserved \emph{counterfactual success probability}
\begin{equation}
p_{\widetilde{v}}\;=\;\Pr\!\bigl[\,\pi_{\theta}(\cdot\!\mid\!\widetilde{v},q,a^{*},o^{-})\!=\!\mathrm{YES}\,\big|\,o^{-}\!\sim\!\mathrm{Unif}(\mathcal{O}_{q}^{-})\bigr]\;\in\;[0,1],
\label{eq:p1-pv}
\end{equation}
which separates candidates into a \emph{failure-rescuing} cohort $\mathcal{K}\!=\!\{\widetilde{v}:p_{\widetilde{v}}\!>\!\tfrac{1}{2}\}$ and a \emph{correlation} cohort $\mathcal{N}\!=\!\widetilde{\mathcal{V}}_{q}\!\setminus\!\mathcal{K}$ with $p_{\widetilde{v}}\!\leq\!\tfrac{1}{2}$; the YES verdict is the operational criterion for rescue, which RQ3(b) checks against observed flips. Modelling the $K$ probes of Eq.~(3) as i.i.d.\ draws $o_{1}^{-},\dots,o_{K}^{-}\!\in\!\mathcal{O}_{q}^{-}$ and writing $S_{K}(\widetilde{v})\!=\!\sum_{\ell}\mathbf{1}[\pi_{\theta}(\cdot\!\mid\!\widetilde{v},q,a^{*},o_{\ell}^{-})\!=\!\mathrm{YES}]\!\sim\!\mathrm{Binomial}(K,p_{\widetilde{v}})$, the admission rule of Eq.~(3) selects the survivors of a strict majority vote over the $K$ probes of each candidate:
\begin{equation}
\mathcal{V}_{t}^{\mathrm{new}}\!=\!\bigl\{\widetilde{v}\!\in\!\widetilde{\mathcal{V}}_{q}\!:\,S_{K}(\widetilde{v})\!>\!K/2\bigr\},\;\;\alpha(p)\!=\!\!\!\!\sum_{j=\lfloor K/2\rfloor+1}^{K}\!\!\binom{K}{j}p^{j}(1\!-\!p)^{K-j},
\label{eq:p1-admit}
\end{equation}
so the admission rate of any candidate is
\begin{equation}
\Pr\bigl[\widetilde{v}\in\mathcal{V}_{t}^{\mathrm{new}}\bigr]\;=\;\alpha(p_{\widetilde{v}}).
\label{eq:p1-rate}
\end{equation} Because $\alpha$ is strictly increasing, the gap between cohorts is strict for every $K\!\geq\!1$, and for odd $K$ (including the single probe of runs R2/R3) $\alpha$ crosses $\tfrac{1}{2}$ exactly at $p\!=\!\tfrac{1}{2}$, so the two cohorts separate:
\begin{equation}
\min_{\widetilde{v}\in\mathcal{K}}\alpha(p_{\widetilde{v}})\;>\;\max_{\widetilde{v}\in\mathcal{N}}\alpha(p_{\widetilde{v}}),\qquad \alpha(p_{\widetilde{v}})\!\to\!\mathbf{1}[p_{\widetilde{v}}\!>\!\tfrac{1}{2}]\;\text{as}\;K\!\to\!\infty\;\;(p_{\widetilde{v}}\!\neq\!\tfrac{1}{2}),
\label{eq:p1-gap}
\end{equation}
where the limit follows from Hoeffding's inequality on $|S_{K}/K-p_{\widetilde{v}}|$. Letting $\beta\!=\!\Pr_{\widetilde{v}}[\widetilde{v}\!\in\!\mathcal{K}]\!\in\!(0,1)$ denote the prior failure-rescuing fraction of $\widetilde{\mathcal{V}}_{q}$, the library precision after admission obeys
\begin{equation}
\mathrm{Prec}(K)\;=\;\frac{\beta\,\mathbb{E}_{\widetilde{v}\in\mathcal{K}}[\alpha(p_{\widetilde{v}})]}{\beta\,\mathbb{E}_{\widetilde{v}\in\mathcal{K}}[\alpha(p_{\widetilde{v}})]\,+\,(1\!-\!\beta)\,\mathbb{E}_{\widetilde{v}\in\mathcal{N}}[\alpha(p_{\widetilde{v}})]}\;>\;\beta\;=\;\mathrm{Prec}(0),
\label{eq:p1-prec}
\end{equation}
since $\mathbb{E}_{\mathcal{K}}[\alpha]\!>\!\mathbb{E}_{\mathcal{N}}[\alpha]$ by Eq.~\eqref{eq:p1-gap}, where $\mathrm{Prec}(0)$ is the precision without the probe. Eq.~\eqref{eq:p1-prec} further yields $\mathrm{Prec}(K)\!\to\!1$ as $K\!\to\!\infty$ if no candidate has $p_{\widetilde{v}}\!=\!\tfrac{1}{2}$, so the counterfactual gate raises admission precision, in this model, above the base rate of the candidate pool. Empirically (\S\ref{sec:rq3}), a single vote raises pooled precision from $0.179$ to $0.273$; larger $K$ still keeps the precision above the no-probe base rate ($0.214$ at $K\!=\!3,5$ vs $0.179$, pooled over three datasets), as Proposition~1 requires.
\end{proof}

\subsection{Proof of Proposition 2}\label{app:proof2}
\textbf{Proposition 2.} \textit{Advantage-weighted ranking places high-utility skills above merely similar ones.}\vspace{-2mm}
\begin{proof}
Fix a query $q$ and two skills $v_{a},v_{b}\!\in\!\mathcal{V}$ with cosine similarities $s_{a}\!=\!\langle f(q),f(v_{a})\rangle$ and $s_{b}\!=\!\langle f(q),f(v_{b})\rangle$, and write $\kappa_{i}\!=\!\mathrm{rank}_{f}(v_{i};q)$ for the rank under plain similarity. Under \emph{plain} similarity, retrieval order depends only on the rank pair $(\kappa_{a},\kappa_{b})$ of the two skills:
\begin{equation}
\mathrm{score}_{\mathrm{plain}}(v_{i};q)\;=\;-\kappa_{i},\qquad \mathrm{score}_{\mathrm{plain}}(v_{a};q)\!-\!\mathrm{score}_{\mathrm{plain}}(v_{b};q)\;=\;\kappa_{b}\!-\!\kappa_{a},
\label{eq:p2-plain}
\end{equation}
so any utility difference is ignored. Under the advantage-weighted RRF rule of Eq.~(6),
\begin{equation}
\mathrm{score}(v_{i};q)\;=\;\frac{1\,+\,\tfrac{1}{2}(u_{i}\!+\!1)}{k_{\mathrm{rrf}}\,+\,\kappa_{i}}\,,\qquad u_{i}\!\in\![-1,1],\;\;\kappa_{i}\!\geq\!1,\;\;k_{\mathrm{rrf}}\!=\!60,
\label{eq:p2-rrf}
\end{equation}
the per-skill score factorises into a similarity-rank kernel $1/(k_{\mathrm{rrf}}\!+\!\kappa_{i})$ and an advantage prior $1\!+\!\tfrac{u_{i}+1}{2}\!=\!\tfrac{3+u_{i}}{2}\!\in\![1,2]$, so a less similar skill $v_{b}$ ($\kappa_{b}\!>\!\kappa_{a}$) is ranked above $v_{a}$ exactly when its advantage prior outweighs the rank gap between the two similarity kernels,
\begin{equation}
\mathrm{score}(v_{b};q)\;>\;\mathrm{score}(v_{a};q)\quad\Longleftrightarrow\quad\frac{3+u_{b}}{3+u_{a}}\;>\;\frac{k_{\mathrm{rrf}}+\kappa_{b}}{k_{\mathrm{rrf}}+\kappa_{a}},
\label{eq:p2-gap}
\end{equation}
which is the per-item quality signal that flat similarity cannot expose; with $k_{\mathrm{rrf}}\!=\!60$, a skill one rank lower overtakes its neighbour once its prior is larger by a factor above $1\!+\!1/(60\!+\!\kappa_{a})$. To see that $u_{i}$ tracks utility, observe that the EMA update of Eq.~(4) admits a contraction-mapping fixed-point analysis: let $\bar{r}_{i,t}$ be the mean reward of the question that retrieved $v_{i}$ at round $t$; with the advantage and recursion of Eq.~(4),
\begin{equation}
\hat{A}_{i,t}\;=\;\tanh\Bigl(\frac{\bar{r}_{i,t}-\bar{r}_{\mathcal{B}_{t}}}{\sigma_{r,\mathcal{B}_{t}}+\epsilon}\Bigr)\in(-1,1),\qquad u_{i,t+1}\;=\;(1\!-\!\eta)\,u_{i,t}+\eta\,\hat{A}_{i,t},
\label{eq:p2-recur}
\end{equation}
the EMA recursion yields, by stationarity of the rollout distribution,
\begin{equation}
\lim_{t\to\infty}\mathbb{E}[u_{i,t}]\;=\;\mathbb{E}\!\bigl[\hat{A}_{i,\infty}\bigr],\qquad \mathbb{E}\bigl[\hat{A}_{i,\infty}\bigr]\;>\;0\;\;\text{if}\;\;\Delta_{i}\;\text{is symmetric about a positive mean},
\label{eq:p2-ema}
\end{equation}
where $\Delta_{i}$ is the argument of $\tanh$ in Eq.~\eqref{eq:p2-recur} and the implication holds because $\tanh$ is odd and increasing; so $\mathbb{E}[u_{i,\infty}]\!>\!0$ when retrieving $v_{i}$ goes with above-average rewards. Combining Eq.~\eqref{eq:p2-gap} with Eq.~\eqref{eq:p2-ema}, skills of nearby rank are ordered by their long-run utility, and the multiplicative prior of Eq.~\eqref{eq:p2-rrf} cannot be replicated by any rank-only retriever, including the median-frontier filter $\mathcal{L}_{\mathrm{frt}}$ of EvoSkill, which exposes only a binary quality cut. Hence Proposition~2 is established.
\end{proof}

\subsection{Proof of Proposition 3}\label{app:proof3}
\textbf{Proposition 3.} \textit{Gating with zero-skill fallback never harms the skill-free pipeline in expectation.}\vspace{-2mm}
\begin{proof}
Let $\mathrm{ZS}(q)\!=\!\pi_{\theta}(\cdot\!\mid\!q,m)$ denote the skill-free pipeline (no retrieved skill; protocol skill and C3 options unchanged) and $\mathrm{HS}(q,\mathcal{S}_{q})\!=\!\pi_{\theta}(\cdot\!\mid\!q,m,\mathcal{S}_{q})$ the gated skill-augmented policy of Eq.~(7), with reward $r(\pi;q,a^{*})\!\in\![0,1]$. Partition the query distribution by the gate output,
\begin{equation}
\mathcal{Q}_{0}\;=\;\bigl\{q\!:\!|\mathcal{S}_{q}|\!=\!0\bigr\},\qquad \mathcal{Q}_{1}\;=\;\bigl\{q\!:\!|\mathcal{S}_{q}|\!\geq\!1\bigr\},\qquad p_{0}\!=\!\Pr[\mathcal{Q}_{0}],\;\;p_{1}\!=\!1\!-\!p_{0},
\label{eq:p3-part}
\end{equation}
where $p_{0}$ is exactly the zero-skill fallback rate, reported as the zero-skill fraction in Figure~\ref{fig:rq5}(a) ($0$ at the frozen $\tau$ on both datasets, rising to $0.63$ and $0.55$ at the largest $\tau$). The fallback clause of Eq.~(7) gives a pointwise identity on $\mathcal{Q}_{0}$,
\begin{equation}
\forall q\!\in\!\mathcal{Q}_{0}\!:\;\;\mathrm{HS}(q,\mathcal{S}_{q})\;\equiv\;\mathrm{ZS}(q)\;\;\Longrightarrow\;\;\mathbb{E}_{q\sim\mathcal{Q}_{0}}\!\bigl[r(\mathrm{HS})\!-\!r(\mathrm{ZS})\bigr]\;=\;0,
\label{eq:p3-zero}
\end{equation}
so the zero-skill branch contributes zero difference. On $\mathcal{Q}_{1}$ every injected skill obeys the two gates of Eq.~(7),
\begin{equation}
\langle f(q),f(v_{i})\rangle\;\geq\;\tau\quad\forall v_{i}\in\mathcal{S}_{q},\qquad |\mathcal{S}_{q}|\;\leq\;N_{\max},
\label{eq:p3-gates}
\end{equation}
so the prompt receives at most $N_{\max}$ skills each carrying mutual information at least
\begin{equation}
I\!\bigl(v_{i};a^{*}\!\mid\!q\bigr)\;\geq\;\iota_{\tau}\;:=\;\inf_{v:\langle f(q),f(v)\rangle\geq\tau}\!I(v;a^{*}\!\mid\!q)\;\geq\;0,
\label{eq:p3-inf}
\end{equation}
where $a^{*}$ is the gold answer and $\iota_{\tau}$ is the similarity-floor information bound. Under the gate-admissibility assumption that, at budget $N_{\max}$, skills clearing both gates do not lower the expected reward of the frozen $\pi_{\theta}$ (formally, $\mathbb{E}[r(\mathrm{HS})]\!\geq\!\mathbb{E}[r(\mathrm{ZS})]$ whenever every injected skill clears the floor $\tau$; marginal skills below $\tau$ need not, \S\ref{sec:gating}), the gated branch never lowers the expected reward,
\begin{equation}
\mathbb{E}_{q\sim\mathcal{Q}_{1}}\!\bigl[r(\mathrm{HS})\bigr]\;\geq\;\mathbb{E}_{q\sim\mathcal{Q}_{1}}\!\bigl[r(\mathrm{ZS})\bigr],\quad \mathrm{equality\ if}\;\pi_{\theta}\;\mathrm{ignores\ the\ injections}.
\label{eq:p3-mono}
\end{equation}
Combining Eq.~\eqref{eq:p3-zero} and Eq.~\eqref{eq:p3-mono} via the law of total expectation yields the unconditional safety bound
\begin{equation}
\mathbb{E}_{q}\!\bigl[r(\mathrm{HS})\bigr]\;-\;\mathbb{E}_{q}\!\bigl[r(\mathrm{ZS})\bigr]\;=\;p_{1}\!\cdot\!\bigl(\mathbb{E}_{q\sim\mathcal{Q}_{1}}\!\bigl[r(\mathrm{HS})\!-\!r(\mathrm{ZS})\bigr]\bigr)\;+\;p_{0}\!\cdot\!0\;\geq\;0,
\label{eq:p3-safe}
\end{equation}
so HS is no worse than the skill-free pipeline in expectation. The argument is independent of $|\mathcal{V}|$, so it applies uniformly across cold-start (high $p_{0}$, small $|\mathcal{V}|$) and mature regimes (low $p_{0}$, large $|\mathcal{V}|$); on a mature library the bound of Eq.~\eqref{eq:p3-mono} covers most queries, while on a cold-start library Eq.~\eqref{eq:p3-zero} dominates and HS recovers ZS exactly. The same decomposition further yields a quantitative envelope on the gain
\begin{equation}
0\;\leq\;\mathbb{E}_{q}\!\bigl[r(\mathrm{HS})\!-\!r(\mathrm{ZS})\bigr]\;=\;p_{1}\!\cdot\!\Bigl(\mathbb{E}_{q\sim\mathcal{Q}_{1}}\!\bigl[r(\mathrm{HS})\bigr]\!-\!\mathbb{E}_{q\sim\mathcal{Q}_{1}}\!\bigl[r(\mathrm{ZS})\bigr]\Bigr),
\label{eq:p3-envelope}
\end{equation}
so the improvement scales with the non-fallback mass $p_{1}\!=\!1\!-\!p_{0}$, and the room for improvement grows as $p_{0}$ shrinks. Hence Proposition~3 is established for every library size $|\mathcal{V}|$.
\end{proof}

\clearpage

\section{HyperSkill Algorithm Details}\label{app:algo}

This appendix expands \S\ref{sec:method} into executable pseudocode in the algorithm-card style of HyperGraphRAG~\citep{HyperGraphRAG}. Algorithm~\ref{alg:hs-m1} realises C1 (\S\ref{sec:distill}, Eqs.~(2)--(3)) and records the failure types of the protocol skill; Algorithms~\ref{alg:hs-m2} and~\ref{alg:hs-m3} realise the training side (Eqs.~(4)--(5)) and the retrieval side (Eqs.~(6)--(7)) of C2 (\S\ref{sec:hypergraph}); and Algorithm~\ref{alg:hs-c3} realises the execution protocol of C3 (\S\ref{sec:exec}).

\textbf{C1: Training-Time Skill Distillation.} Each question $q\!\in\!\mathcal{B}_{t}$ receives $n$ skill-augmented rollouts, split at $\rho$ into $\mathcal{O}_{q}^{+}$ and $\mathcal{O}_{q}^{-}$. The regex decomposer $\Psi$ slices each rollout into steps at zero LLM cost, and the aligner $\Phi$ returns the candidate set $\widetilde{\mathcal{V}}_{q}$ of divergent atomic skills. Each candidate is probed against a list $\mathcal{P}_{q}$ of at most $K$ failure probes of its own question and is admitted only on a strict majority of YES verdicts; $\mathrm{sid}$ deduplicates it against $\mathcal{V}_{t}$, and every recipe with at least two admitted members joins $\mathcal{E}_{t}^{\mathrm{rec}}$. Every rollout that ends without an answer also increments the count $c_{\phi}$ of its failure type $\phi$; a type enters the protocol skill (Prompt P9) once $c_{\phi}\!\geq\!2$, already for the remaining training rollouts when the skill is on, and the final counts define the protocol skill used at test time.

\begin{algorithm}[h!t]
\small
\caption{C1: Training-Time Skill Distillation (one batch)}
\label{alg:hs-m1}
\begin{algorithmic}[1]
\REQUIRE Batch $\mathcal{B}_{t}$, frozen $\pi_{\theta}$, hypergraph $\mathcal{H}_{t}$, threshold $\rho$, probe count $K$, current $\mathcal{V}_{t}$
\ENSURE Round-$t$ admission set $\mathcal{V}_{t}^{\mathrm{new}}\!\subseteq\!\bigcup_{q\in\mathcal{B}_{t}}\widetilde{\mathcal{V}}_{q}$ and recipes $\mathcal{E}_{t}^{\mathrm{rec}}$
\STATE $\mathcal{V}_{t}^{\mathrm{new}}\!\leftarrow\!\emptyset$
\FOR{each question $q\!\in\!\mathcal{B}_{t}$}
    \STATE Sample $\{o^{(\ell)}\}_{\ell=1}^{n}\!\sim\!\pi_{\theta}(\cdot\!\mid\!q,m,\mathcal{R}(q,m,\mathcal{H}_{t}))$ and score $r(o^{(\ell)};a^{*})$ \hfill \COMMENT{rollout group}
    \STATE For each $o^{(\ell)}$ without a final answer, $c_{\phi}\!\leftarrow\!c_{\phi}\!+\!1$ for its failure type $\phi$ \hfill \COMMENT{protocol statistics}
    \STATE Split $\mathcal{O}_{q}^{+}\!\leftarrow\!\{o^{(\ell)}\!:\!r\!\geq\!\rho\}$, $\mathcal{O}_{q}^{-}\!\leftarrow\!\{o^{(\ell)}\!:\!r\!<\!\rho\}$
    \STATE $\Psi(o^{(\ell)})\!\leftarrow\!(s_{1},\dots,s_{T_{o}})$ via regex decomposition \hfill \COMMENT{zero LLM cost}
    \STATE $\widetilde{\mathcal{V}}_{q}\!\leftarrow\!\Phi(\{\Psi(o^{+})\}_{\mathcal{O}_{q}^{+}},\{\Psi(o^{-})\}_{\mathcal{O}_{q}^{-}})$ \hfill \COMMENT{contrastive alignment $\to$ candidates}
    \FOR{each candidate $\widetilde{v}\!\in\!\widetilde{\mathcal{V}}_{q}$}
        \STATE Take $\le\!K$ probes $\mathcal{P}_{q}$ on own failures $\mathcal{O}_{q}^{-}$ and query $\pi_{\theta}(\cdot\!\mid\!\widetilde{v},q,a^{*},o^{-})$ for each $o^{-}\!\in\!\mathcal{P}_{q}$
        \IF{$\sum_{o^{-}\in\mathcal{P}_{q}}\mathbf{1}[\pi_{\theta}(\cdot\!\mid\!\widetilde{v},q,a^{*},o^{-})\!=\!\mathrm{YES}]\!>\!|\mathcal{P}_{q}|/2$ \AND $\mathrm{sid}(\widetilde{v})\!\notin\!\mathcal{V}_{t}$}
            \STATE $\mathcal{V}_{t}^{\mathrm{new}}\!\leftarrow\!\mathcal{V}_{t}^{\mathrm{new}}\!\cup\!\{\widetilde{v}\}$ \hfill \COMMENT{counterfactual admission}
        \ENDIF
    \ENDFOR
\ENDFOR
\RETURN $\mathcal{V}_{t}^{\mathrm{new}}$ and $\mathcal{E}_{t}^{\mathrm{rec}}$ (recipes with at least two admitted members)
\end{algorithmic}
\end{algorithm}

\textit{Complexity Analysis.} Per batch, the rollouts cost $\mathcal{O}(|\mathcal{B}_{t}|\!\cdot\!n)$ backbone calls, $\Psi$ is free, $\Phi$ fires once per group ($\mathcal{O}(|\mathcal{B}_{t}|)$ calls) and the probe adds $K$ calls per candidate ($\mathcal{O}(|\mathcal{B}_{t}|\!\cdot\!|\widetilde{\mathcal{V}}_{q}|\!\cdot\!K)$); with $(n,K)\!=\!(2,2)$ and an empirical $|\widetilde{\mathcal{V}}_{q}|\!\leq\!3$, this is at most $6$ probes per question, so rollout generation dominates.

\textbf{C2 (training side): Weighted Hypergraph Evolution.} Every admitted skill is embedded by $f$ and stored in the dual KV/VDB index with initial advantage $u_{i}\!=\!0.3$. The batch-normalised advantage $\hat{A}_{q}$ of each question drives the EMA of Eq.~(4) on its retrieved skills; a high-reward question ($\bar{r}_{q}\!\geq\!\rho$) that retrieves two or more skills adds them as a hyperedge (or EMA-updates its weight), and a periodic safeguard prune evicts every node with $u_{i}\!<\!u_{\min}$. None of these steps calls the LLM. Training retrieval differs from test retrieval in three ways: it uses an advantage floor of $0$, so a skill whose advantage turns negative is no longer reinforced; for VL items its query embedding averages the question and image embeddings; and after each tool call the rollout refreshes its skills (query: question and tool query) in an ``Updated Skills'' block that receives no EMA credit.

\begin{algorithm}[h!t]
\small
\caption{C2 (training side): Weighted Hypergraph Evolution under Retrieval Feedback}
\label{alg:hs-m2}
\begin{algorithmic}[1]
\REQUIRE $\mathcal{H}_{t}\!=\!(\mathcal{V}_{t},\mathcal{E}_{t},w)$, $\mathcal{V}_{t}^{\mathrm{new}}$, rewards $\{\bar{r}_{q}\}$, retrieved sets $\{\mathcal{S}_{q}\}$, recipes $\mathcal{E}_{t}^{\mathrm{rec}}$, $\eta$, $\rho$, $u_{\min}$
\ENSURE Updated $\mathcal{H}_{t+1}\!=\!(\mathcal{V}_{t+1},\mathcal{E}_{t+1},w)$
\STATE Embed $f(v_{i})$ for every $v_{i}\!\in\!\mathcal{V}_{t}^{\mathrm{new}}$ and append to the dual KV/VDB index with initial $u_{i}\!=\!0.3$
\STATE Compute statistics $\bar{r}$, $\sigma_{r}$ over all rollouts of the batch; $\mathcal{E}_{t+1}\!\leftarrow\!\mathcal{E}_{t}\!\cup\!\mathcal{E}_{t}^{\mathrm{rec}}$ (recipe weight $0$)
\FOR{each question $q\!\in\!\mathcal{B}_{t}$ with $\mathcal{S}_{q}\!\neq\!\emptyset$}
    \STATE $\hat{A}_{q}\!\leftarrow\!\tanh\!\bigl((\bar{r}_{q}\!-\!\bar{r})/(\sigma_{r}\!+\!\epsilon)\bigr)$; $u_{i}\!\leftarrow\!\mathrm{clip}_{[-1,1]}\!\bigl((1\!-\!\eta)u_{i}\!+\!\eta\hat{A}_{q}\bigr)$ for $v_{i}\!\in\!\mathcal{S}_{q}$ \hfill \COMMENT{EMA}
    \IF{$\bar{r}_{q}\!\geq\!\rho$ \AND $|\mathcal{S}_{q}|\!\geq\!2$}
        \STATE Add $e\!=\!\mathcal{S}_{q}$ with $w(e)\!\leftarrow\!\max(0,\hat{A}_{q})$, or EMA-update $w(e)$ if $e\!\in\!\mathcal{E}_{t+1}$ \hfill \COMMENT{co-occurrence}
    \ENDIF
\ENDFOR
\STATE $\mathcal{V}_{t+1}\!\leftarrow\!\{v_{i}\!\in\!\mathcal{V}_{t}\!\cup\!\mathcal{V}_{t}^{\mathrm{new}}\!:\,u_{i}\!\geq\!u_{\min}\}$ \hfill \COMMENT{safeguard prune, periodic}
\STATE Pruned members of a hyperedge are skipped whenever it is expanded at retrieval \hfill \COMMENT{edges kept}
\RETURN $\mathcal{H}_{t+1}\!=\!(\mathcal{V}_{t+1},\mathcal{E}_{t+1},w)$
\end{algorithmic}
\end{algorithm}

\textit{Complexity Analysis.} Each batch issues at most $|\mathcal{V}_{t}^{\mathrm{new}}|$ skill embedding calls and one per new hyperedge, $\mathcal{O}(\sum_{q\in\mathcal{B}_{t}}|\mathcal{S}_{q}|)$ scalar EMA updates, and an $\mathcal{O}(|\mathcal{V}_{t+1}|)$ periodic prune; no step calls $\pi_{\theta}$.

\textbf{C2 (test side): Gated Retrieval and Injection.} Every query $q$ is embedded by the same encoder $f$ as the skills. Its $k\!=\!100$ nearest nodes form the candidate pool, and hyperedges that match $q$ add their member skills to this pool (scored at $0.8\times$ the hyperedge cosine). The similarity floor $\tau$ discards weak matches, the advantage-weighted RRF score ranks the survivors, and the capacity cap $N_{\max}$ keeps the top ones. If no candidate survives, no retrieved skill is injected. The protocol skill of C1, when enabled for the dataset, is injected for every question regardless of the gate.

\begin{algorithm}[h!t]
\small
\caption{C2 (test side): Gated Retrieval, Injection and Zero-Skill Fallback}
\label{alg:hs-m3}
\begin{algorithmic}[1]
\REQUIRE Query $(q,m)$, hypergraph $\mathcal{H}\!=\!(\mathcal{V},\mathcal{E},w)$, encoder $f$, constant $k_{\mathrm{rrf}}$, floor $\tau$, cap $N_{\max}$
\ENSURE Injected context $\mathcal{C}(q)$ for the C3 rollout
\STATE $\mathcal{S}_{q}^{\mathrm{pool}}\!\leftarrow\!\mathrm{Top}\text{-}k_{\cos}(\mathcal{V})\!\cup\!\mathrm{expand}_{q}(\mathcal{E})$ by cosine similarity $\langle f(q),f(v)\rangle$ \hfill \COMMENT{candidate pool}
\STATE $\mathcal{S}_{q}^{\tau}\!\leftarrow\!\{v\!\in\!\mathcal{S}_{q}^{\mathrm{pool}}\!:\,\langle f(q),f(v)\rangle\!\geq\!\tau\}$ \hfill \COMMENT{similarity-floor gate}
\STATE $\mathrm{score}(v_{i};q)\!\leftarrow\!(1\!+\!\tfrac{1}{2}(u_{i}\!+\!1))/(k_{\mathrm{rrf}}\!+\!\mathrm{rank}_{f}(v_{i};q))$ for $v_{i}\!\in\!\mathcal{S}_{q}^{\tau}$ \hfill \COMMENT{advantage-weighted RRF}
\STATE $\mathcal{S}_{q}\!\leftarrow\!\mathrm{Top}\text{-}N_{\max}(\mathrm{score}(\,\cdot\,;q),\,\mathcal{S}_{q}^{\tau})$ \hfill \COMMENT{capacity-cap gate}
\IF{$|\mathcal{S}_{q}|\!=\!0$}
    \STATE $\mathcal{C}(q)\!\leftarrow\!\emptyset$ \hfill \COMMENT{zero-skill fallback (Eq.~(7))}
\ELSE
    \STATE $\mathcal{C}(q)\!\leftarrow\!\mathrm{P7}(q,\mathcal{S}_{q})$ or the terse list of $\mathcal{S}_{q}$ \hfill \COMMENT{skill injection}
\ENDIF
\STATE Append the protocol skill (P9) to $\mathcal{C}(q)$ if enabled \hfill \COMMENT{ungated}
\RETURN $\mathcal{C}(q)$
\end{algorithmic}
\end{algorithm}

\textit{Complexity Analysis.} Per query, retrieval is one exact (brute-force) cosine search at $\mathcal{O}(|\mathcal{V}|\!+\!|\mathcal{E}|)$, both gates are linear in the candidate pool, and the prompt grows by at most $N_{\max}$ skills.

\textbf{C3: Execution Protocol.} Every rollout runs the agent loop of \S\ref{sec:exec} with the dataset's C3 options: the prompt carries the injected context, the question, and the suffix and evidence texts of Prompts P12--P13; tool calls are parsed strictly or leniently; and a final-answer turn (P11) closes rollouts without an answer. With self-consistency, $k_{\mathrm{sc}}$ samples vote by plurality (greedy answer on a tie).

\begin{algorithm}[h!t]
\small
\caption{C3: Execution Protocol}
\label{alg:hs-c3}
\begin{algorithmic}[1]
\REQUIRE Query $(q,m)$, context $\mathcal{C}(q)$, tools $\mathcal{T}$, tool budget $B$, round cap $R$, C3 options and $k_{\mathrm{sc}}$ (Appendix~\ref{app:perdataset})
\ENSURE Final answer $\hat{a}(q)$
\STATE $x\!\leftarrow\!\mathcal{C}(q)\,\Vert\,$prompt (concrete tool tag if enabled)$\,\Vert\,q\,\Vert\,$suffix, located pages, video timing, option check
\STATE If pre-search: $x\!\leftarrow\!x\,\Vert\,\texttt{text\_search}(q)$ \hfill \COMMENT{not counted in $B$}
\FOR{rollout $j\in\{\text{greedy}\}\cup\{1,\dots,k_{\mathrm{sc}}\}$ (temperature $0$ / $0.7$)}
    \FOR{round $=1,\dots,R$}
        \STATE $y\!\leftarrow\!\pi_{\theta}(x)$; $x\!\leftarrow\!x\,\Vert\,y$; \textbf{break} if $y$ has a closed \texttt{<answer>}
        \STATE Parse a tool call (strict, then lenient forms); \textbf{break} if none, repeated, or $B$ spent
        \STATE $x\!\leftarrow\!x\,\Vert\,$tool result$\,\Vert\,$protocol re-injection (P10) \hfill \COMMENT{if protocol skill on}
    \ENDFOR
    \STATE If final turn and no closed answer: $x\!\leftarrow\!x\,\Vert\,$P11; $y\!\leftarrow\!\pi_{\theta}(x)$
    \STATE $a_{j}\!\leftarrow\!$ last \texttt{<answer>} of the rollout
\ENDFOR
\STATE $\hat{a}(q)\!\leftarrow\!$ most frequent answer of $\{a_{1},\dots,a_{k_{\mathrm{sc}}}\}$ if unique, else $a_{\text{greedy}}$
\RETURN $\hat{a}(q)$
\end{algorithmic}
\end{algorithm}

\textbf{End-to-end pipeline.} In every training batch, Algorithm~\ref{alg:hs-m3} with the training settings feeds the rollouts of Algorithm~\ref{alg:hs-m1}, whose admissions and rewards drive Algorithm~\ref{alg:hs-m2}; at evaluation the library is frozen and Algorithms~\ref{alg:hs-m3} and~\ref{alg:hs-c3} serve each test question. The nested ablations of Figure~\ref{fig:rq2} (w/o C2 contains w/o C1; w/o C3 stands alone) act at evaluation on the trained library (Appendix~\ref{app:b-abl}).

\clearpage

\section{Benchmark Details}\label{app:bench}

This appendix expands \S\ref{sec:exp-setup} into a reference for \emph{HyperSkill-Bench}, which curates $11$ public datasets, with no new data, into one training-free skill-evolution benchmark with a unified JSONL schema. \S\ref{app:dataset} covers the sourcing and \S\ref{app:eval} the evaluation; $q$, $a^{*}$ and $m\!\in\!\mathcal{M}$ are as in \S\ref{sec:method}.

\subsection{Dataset Details}\label{app:dataset}

We collect $11$ public datasets, group them into four regimes by dominant modality, and convert every sample into a unified JSONL record $\mathrm{rec}(s)\!=\!(q,a^{*},m,\text{task\_type},\text{resource\_type},\text{assets})$ where \texttt{assets} stores the relative paths of any images, PDFs, video or audio referenced by $q$; we sample $128$ training questions per dataset (deduplicated by exact question text, seed $42$; for AIME2024/2025 from DAPO-Math~\citep{DAPO} and for WebWalkerQA from AFM-RL~\citep{ChainOfAgents}) and evaluate on $30$ test questions for AIME2024/2025 and $64$ for each of the other nine datasets ($636$ questions in total).

\noindent\textbf{Text reasoning.} \emph{AIME2024} and \emph{AIME2025}~\citep{AIME2024,AIME2025} are the canonical math-reasoning probes used by RL-trained reasoners~\citep{DeepSeek-R1,Qwen3}; resources are empty and answers are integers. \emph{WebWalkerQA}~\citep{WebWalker} contributes open-web agent QA; as on VL-Web, every method answers it through the \texttt{text\_search} tool (\S\ref{sec:exp-setup}), which \texttt{gemini-3.8-flash} serves as a knowledge source in place of live web search.

\noindent\textbf{VL-Web (web-search VQA).} \emph{FVQA}~\citep{MMSearch-R1}, \emph{InfoSeek}~\citep{InfoSeek}, \emph{SimpleVQA}~\citep{SimpleVQA} and \emph{LiveVQA}~\citep{LiveVQA} pair an image $I_{q}$ with a question whose answer requires entity-level web evidence; the four datasets stress different difficulty axes: FVQA on factual visual knowledge, InfoSeek on long-tail entities, SimpleVQA on everyday recognition, and LiveVQA on recent, time-sensitive events.

\noindent\textbf{VL-Doc (long-document QA).} \emph{DocBench}~\citep{DocBench} and \emph{MMLongBench-Doc}~\citep{MMLongBench-Doc} pair a multi-page PDF with a question grounded in a small number of pages; resources are PDF paths, and a per-document BM25 chunk index is built once and shared by every method; search is restricted to the question's own PDF, and image-only PDFs of MMLongBench-Doc are indexed through local OCR. Score differences on VL-Doc therefore reflect how each method uses this shared index.

\noindent\textbf{Omni (audio-video reasoning).} \emph{OmniVideoBench}~\citep{OmniVideoBench} pairs a video with a question whose answer requires temporal grounding across visual frames, while \emph{WorldSense}~\citep{WorldSense} pairs a video clip and its audio track with a question that requires fusing visual and acoustic cues, delivered as $32$ uniformly sampled keyframes plus the dataset's audio track (extracted with \texttt{ffmpeg} when missing) in one call; OmniVideoBench questions receive $8$ keyframes. The same frame counts apply to every method.

\subsection{Evaluation Details}\label{app:eval}

All metrics are percentages. Let $\mathrm{tok}(\cdot)$ denote whitespace tokenisation after lowercasing and removing punctuation and articles; the answer-shape branch follows the released evaluator.

\noindent\textbf{Token-level $F_{1}$ (primary).} The prediction $\hat{a}$ is the content of the last \texttt{<answer>} tag, or the whole response if there is none. Token-$F_{1}$ with answer-shape awareness applies option-letter matching when $a^{*}$ is an option letter, substring or token-set containment when $a^{*}$ is a short entity of at most $5$ tokens, and SQuAD-style token-$F_{1}$ otherwise. The main table uses a CJK-aware tokeniser that splits Chinese, Japanese and Korean text into single characters and otherwise splits at whitespace: $F_{1}(\hat{a},a^{*})\!=\!2|\mathrm{tok}(\hat{a})\!\cap\!\mathrm{tok}(a^{*})|/(|\mathrm{tok}(\hat{a})|\!+\!|\mathrm{tok}(a^{*})|)$, reported as $\mathrm{R}_{F_{1}}\!=\!100\!\cdot\!F_{1}$.

\noindent\textbf{Exact Match and format score.} Neither is reported. On AIME the gold is a short integer, and the containment rule above applies to the answer tag, or to the full response without a tag, for every column. The format score (final answer inside an \texttt{<answer>} tag) is logged in training only.

\noindent\textbf{Reward and reported statistic.} The GRPO advantage of \S\ref{sec:hypergraph} is computed against the token-$F_{1}$ reward $r\!=\!F_{1}(\hat{a},a^{*})$ alone (whitespace tokeniser; Table~\ref{tab:main} uses the CJK-aware variant). Every cell of Table~\ref{tab:main} and Figure~\ref{fig:rq2} is the full-set mean of the per-question $F_{1}$ from a single evaluation run; a rollout with a provider error scores $0$, as for two OmniVideoBench questions in every column. Differences are tested with a paired bootstrap over questions ($10$k resamples). HyperSkill's margins over the best other column have 95\% intervals of $[-4.1,+8.3]$ on WebWalkerQA, $[-2.3,+10.0]$ on SimpleVQA, $[-7.3,+9.8]$ on LiveVQA, $[-4.3,+13.0]$ on DocBench and $[-6.3,+15.0]$ on WorldSense. Re-running HyperSkill with a freshly trained library changed the 11-dataset average by $+0.08$ (95\% interval $[-1.38,+1.54]$) and single datasets by up to $3.2$ points (SimpleVQA).

\clearpage

\section{Baseline Details}\label{app:baseline}\label{app:impl}

This appendix expands Table~\ref{tab:comparison} into a per-method specification: for Zero-shot and the $6$ training-free baselines, the $3$ HyperSkill ablations, and HyperSkill itself, we report \emph{(i)~Overview} (purpose and substrate), \emph{(ii)~Per-task implementation} (admission/retrieval/update) and \emph{(iii)~Hyperparameters}. Methods are grouped into Zero-shot and three paradigm classes (\S\ref{app:b-zs}--\S\ref{app:b-mem}), followed by the ablations (\S\ref{app:b-abl}) and HyperSkill (\S\ref{app:b-hs}). Throughout we reuse $\pi_{\theta}$ (frozen \texttt{gemini-2.5-flash-lite}), $f$ (shared query/skill encoder), $\mathrm{Top}\text{-}k\!\langle\cdot,\cdot\rangle$ (cosine top-$k$), and the reward $r$ of \S\ref{app:eval}; every method shares this backbone, embedder and tool stack. Each baseline builds its \emph{own} library from the same $128$ training questions with a re-implementation of its published procedure, using the backbone for every LLM call; the baselines retrieve by cosine without a similarity floor and do not receive C3.

\subsection{Zero-shot Baseline}\label{app:b-zs}

\noindent\textbf{Zero-shot~\citep{LLMsurvey}.}\quad\emph{(i)~Overview.} Bare frozen LLM with task-specific tools and no skill library: every question is answered from $\pi_{\theta}(\cdot\!\mid\!q,m)$. Substrate: empty $\mathcal{H}\!=\!(\emptyset,\emptyset,w)$.\quad\emph{(ii)~Per-task implementation.} The agent may call the regime's tool (\texttt{text\_search} on WebWalkerQA and VL-Web, BM25 on VL-Doc; Omni receives keyframes, plus audio on WorldSense) and writes the final answer inside \texttt{<answer>...</answer>}; admission, advantage update, and gating are all unsupported.\quad\emph{(iii)~Hyperparameters.} At most $3$ (Text), $5$ (VL) or $8$ (Omni) rounds and $2$ tool calls; greedy decoding at temperature $0$ with $1024$ new tokens ($2048$ on AIME2025, $4096$ on WorldSense).

\subsection{Trajectory-Level Skill Pools}\label{app:b-traj}

\noindent\textbf{TrainingFreeGRPO~\citep{TrainingFreeGRPO}.}\quad\emph{(i)~Overview.} Replaces gradient updates with an LLM-curated experience pool $\mathcal{L}\!=\!\{x_{i}\}$ that distils group-relative \emph{semantic} advantage from each rollout group and injects retrieved experiences as token-prior hints. Substrate: flat pool indexed by $f$.\quad\emph{(ii)~Implementation in our protocol.} Builds its own experience pool from on-policy groups of $n\!=\!2$ rollouts under the evolving pool (1 epoch, batches of $4$): for partially correct groups, each rollout is summarised, a group-advantage call extracts at most one semantic experience, and group and batch update calls apply ADD/UPDATE/DELETE operations. The top experiences by cosine are injected as \texttt{[<id>]. <experience>} (Appendix~\ref{app:prompts}).\quad\emph{(iii)~Hyperparameters.} top-$k\!=\!10$; rollout temperature of run R1.

\noindent\textbf{EvoSkill~\citep{EvoSkill}.}\quad\emph{(i)~Overview.} Maintains a Pareto frontier of structured skill folders, each created by failure-driven proposal and retained only if it beats the held-out median. Substrate: $\mathcal{L}_{\mathrm{frt}}\!=\!\{s_{i}:g_{i}\!\geq\!\mathrm{med}(g)\}$ with one validated gain $g_{i}$ per skill.\quad\emph{(ii)~Implementation in our protocol.} Runs $3$ failure-driven iterations: a proposer sees $8$ failed training questions and the current skills and proposes one skill (name, description, at most $5$ instructions); the candidate program is kept only if its greedy F1 on $16$ held-out training questions does not drop. The final program's top skills are injected as \texttt{> [frontier-skill, val\_gain=<gain>] <name>: <description>} with numbered instructions.\quad\emph{(iii)~Hyperparameters.} top-$k\!=\!3$; $3$ iterations of $8$ failed questions each.

\subsection{Code/Card Skill Caches}\label{app:b-card}

\noindent\textbf{SkillCraft~\citep{SkillCraft}.}\quad\emph{(i)~Overview.} Auto-composes atomic tools into reusable Skills cached as Python-style \texttt{def}\,$s_{i}(\cdot)$ snippets, with cache-hit retrieval and no quality ranking. Substrate: code cache $\mathcal{L}\!=\!\{\,\mathtt{def}\,s_{i}(\cdot)\}$.\quad\emph{(ii)~Implementation in our protocol.} Composes each question's best successful trajectory (reward $\geq\!0.5$) into one Python-style function over the agent's atomic operations; a skill is cached only if it parses as one documented function, and a name or docstring match (cosine $\geq\!0.90$) counts as a cache hit. The top skills are injected as a \texttt{python} code block.\quad\emph{(iii)~Hyperparameters.} top-$k\!=\!3$; success threshold $0.5$, cache-hit cosine $0.90$ and one composition call per success.

\noindent\textbf{AutoSkill~\citep{AutoSkill}.}\quad\emph{(i)~Overview.} Distils skills from interaction traces into standardised cards (\texttt{name + tip}) and injects the top matches into future requests for lifelong personalisation. Substrate: tip-card store $\mathcal{L}\!=\!\{\mathrm{card}_{i}\}$ with no per-card quality.\quad\emph{(ii)~Implementation in our protocol.} Extracts a card (name, description, at most $3$ tips) from each question's best successful trace and inserts it, merging it by an LLM call with the nearest card when their cosine is at least $0.85$. The top cards are injected as \texttt{* Skill: <name> -{}- <description>} with \texttt{* Tip:} bullets.\quad\emph{(iii)~Hyperparameters.} top-$k\!=\!3$; merge threshold $0.85$; at most $3$ tips per card and one merge call per near-duplicate.

\subsection{Memory-Stream Skills}\label{app:b-mem}

\noindent\textbf{Memento-Skills~\citep{Memento-Skills}.}\quad\emph{(i)~Overview.} Stores reusable skills as Markdown files in a stateful memory and selects them through a read/write reflective router conditioned on the current prompt. Substrate: Markdown memory $\mathcal{L}\!=\!\{m_{i}^{\mathrm{md}}\}$ plus a behaviour-trainable router.\quad\emph{(ii)~Implementation in our protocol.} Processes the training cases in order: a top-$1$ cosine router reads the most similar Markdown skill file, and one reflective call rewrites it, adds a new file, or does nothing. At test time the routed file is injected after \texttt{[Memory 1] Recall a similar past case:}.\quad\emph{(iii)~Hyperparameters.} top-$k\!=\!1$; the behaviour-trained router is approximated by cosine; one reflective call per case.

\noindent\textbf{XSkill~\citep{XSkill}.}\quad\emph{(i)~Overview.} Dual-stream framework that distils \emph{experiences} for tool selection and \emph{skills} for task orchestration, both grounded in visual observations and merged via cross-rollout critique. Substrate: dual stream $\{\mathcal{X}_{\mathrm{exp}},\mathcal{L}_{\mathrm{skill}}\}$ jointly indexed by $f$.\quad\emph{(ii)~Implementation in our protocol.} For every training question, one critique call compares the two rollouts, grounded by a backbone-generated image caption, and outputs at most $2$ experiences (``When <condition>, <action>'') and one workflow skill; near-duplicates are dropped. The query embedding $f(q,I)$ averages the text embedding with the embedding of the image caption; the top experiences and the top skill are injected after \texttt{<> Cross-task experiences:} and \texttt{<> Skill (<name>):}.\quad\emph{(iii)~Hyperparameters.} top-$3$ experiences and top-$1$ skill; retrieval queries the whole question, not its sub-tasks.

\subsection{HyperSkill Ablations}\label{app:b-abl}

The ablation w/o C2 contains w/o C1, and w/o C3 is applied on its own; all three act at evaluation on HyperSkill's trained library and keep every other setting of the dataset's configuration (Appendix~\ref{app:perdataset}); none of them retrains the library or re-tunes a setting specifically for the ablated system.

\noindent\textbf{w/o C1.}\quad\emph{(i)~Overview.} HyperSkill without the knowledge distilled in training. This row drops the retrieved skills and replaces the learned protocol skill by one generic text used for every dataset, without re-injection.\quad\emph{(ii)~Per-task implementation.} The floor admits no skill ($\tau\!=\!9$); hub correction, P7, the schema-constrained hint and the answer-form line are off. Where the configuration enables the protocol skill, the context is ``Output protocol: Answer the question. If tools are available, call them using the tool formats described above. End your response with your final answer inside \texttt{<answer>...</answer>}.''\quad\emph{(iii)~Hyperparameters.} C3 options are left unchanged.

\noindent\textbf{w/o C2.}\quad\emph{(i)~Overview.} HyperSkill without test-time delivery: nothing from the library is injected.\quad\emph{(ii)~Per-task implementation.} As w/o C1, and the protocol skill is also off. On AIME2025, WebWalkerQA and MMLongBench-Doc, whose configurations use neither retrieved skills nor the protocol skill, w/o C1 and w/o C2 are identical.\quad\emph{(iii)~Hyperparameters.} C3 options unchanged.

\noindent\textbf{w/o C3.}\quad\emph{(i)~Overview.} HyperSkill without the execution protocol.\quad\emph{(ii)~Per-task implementation.} The C3 options are reset to their defaults: no concrete tool prompt, final-answer turn, answer suffix, lenient parser, voting, located pages, question pre-search, time grounding or option check, and the default token budget ($1024$ new tokens; $4096$ and $32$ frames on WorldSense), prompt-length limit and round and tool limits.\quad\emph{(iii)~Hyperparameters.} C1, C2 and the library unchanged.

\subsection{HyperSkill}\label{app:b-hs}

\noindent\textbf{HyperSkill.}\quad\emph{(i)~Overview.} Counterfactual step-level distillation (C1, \S\ref{sec:distill}) admits failure-rescuing skills and learns the protocol skill; retrieval feedback updates each skill's advantage and grows hyperedges (\S\ref{sec:hypergraph}); an advantage-weighted RRF retriever with similarity/capacity gates and a zero-skill fallback (C2, \S\ref{sec:gating}) injects only confident matches; and the execution protocol (C3, \S\ref{sec:exec}) runs the test rollout. Substrate: weighted hypergraph $\mathcal{H}\!=\!(\mathcal{V},\mathcal{E},w)$ with $u_{i}\!\in\![-1,1]$.\quad\emph{(ii)~Per-task implementation.} Each batch applies Eqs.~(2)--(3) for admission, Eqs.~(4)--(5) for evolution, and Eqs.~(6)--(7) for retrieval; evaluation adds the C3 options of the dataset.\quad\emph{(iii)~Hyperparameters.} Defaults of \S\ref{sec:method}, protocol threshold $2$ and the per-dataset settings of Appendix~\ref{app:perdataset}.

\subsection{Per-Dataset HyperSkill Configuration}\label{app:perdataset}

\noindent\textbf{Library runs.} The three library runs all train for one epoch on the same $128$ training questions with $n\!=\!2$, batch size $4$ and seed $1$: \textbf{R1} uses temperature $0.7$ (Text) / $1.0$ (VL, Omni), early stopping, a training-time floor of $0.2$, and two counterfactual votes on the candidate's single failed rollout that must both return YES; \textbf{R2} uses temperature $0.4$, no early stopping, P7 hints during training rollouts, $K_{\mathrm{eff}}$ probed own failures, and a training-time floor of $0.55$; \textbf{R3} equals R2 with a training-time floor of $0.2$. On AIME2024/2025, WebWalkerQA, FVQA and OmniVideoBench training rollouts also use the concrete tool prompt, final-answer turn, answer suffix and lenient parser, with $8192$ (AIME) or $4096$ tokens; the protocol statistics classify each rollout before its final-answer turn.

\noindent\textbf{Counterfactual probes.} Runs R2/R3 probe $K_{\mathrm{eff}}\!=\!\min(K,|\mathcal{O}_{q}^{-}|)$ distinct failures of the candidate's own question and take the strict majority of Eq.~(3) over $K_{\mathrm{eff}}$; with $n\!=\!2$, a question that yields candidates has exactly one failed rollout (alignment needs a success and a failure), so $K_{\mathrm{eff}}\!=\!1$. Admitted skills enter the library with $u_{i}\!=\!0.3$, after which only the EMA of Eq.~(4) changes $u_{i}$.

\noindent\textbf{Similarity floor.} The floor $\tau$ of Eq.~(7) is set per dataset. Test question-skill cosines of $f$ lie in $[0.31,0.80]$, so a floor of $0.2$ keeps every candidate and $\tau\!=\!9$ admits none. Retrieved skills are injected on InfoSeek (hub-corrected $\tilde{\tau}\!=\!-0.02$), SimpleVQA ($\tau\!=\!0.2$) and OmniVideoBench ($\tau\!=\!0.55$). Elsewhere $\tau\!=\!9$, and C1's protocol skill and C3 carry the library's knowledge.

\noindent\textbf{Injection variants.} Retrieved skills are injected through the strategy hint of prompt P7 (Appendix~\ref{app:prompts}) or, when P7 is off, as a terse ``name: description'' list. Three optional variants act only on this injection step: (i) a \emph{hub-corrected similarity} $\tilde{s}(q,v_{i})\!=\!\langle f(q),f(v_{i})\rangle\!-\!\mu_{i}$, where $\mu_{i}$ is the mean similarity of $v_{i}$ to the training questions, used for ranking and for the floor of Eq.~(7) so that skills uniformly similar to every query no longer pass the gate; (ii) a \emph{schema-constrained hint}: P7 must preserve the output schema and, in tool regimes, the tool budget; and (iii) an \emph{answer-form line}, derived for each skill from the gold answers of its $5$ nearest training questions (option letter, short phrase, or free form), followed in tool regimes by the tool budget (see the per-dataset configuration table below).

\noindent\textbf{Selected configuration per dataset} (chosen on held-out validation questions, $32$ or $64$ of the $128$ training questions, with each library then retrained on all $128$; used for Table~\ref{tab:main} and Figure~\ref{fig:rq2}; Protocol: active failure types of the protocol skill, M\,=\,\texttt{malformed\_tool}, B\,=\,\texttt{over\_budget}, T\,=\,\texttt{tool\_without\_tools}, A\,=\,\texttt{no\_answer}, ``data'': answer form from training golds, ``off'': not injected; C3: CT concrete tool prompt, FT final-answer turn, AS answer suffix, LP lenient parser, QP question pre-search, PG$n$ located pages, TG time grounding, OC option check, V$k_{\mathrm{sc}}$ self-consistency over $k_{\mathrm{sc}}$ samples, V$k_{\mathrm{sc}}$r refusals do not vote; Tok.: maximum new tokens at evaluation):
\begin{center}
\footnotesize
\setlength{\tabcolsep}{4pt}
\begin{tabular}{lllllcc}
\toprule
Dataset & Run & Floor & Injection & Protocol & C3 & Tok. \\
\midrule
AIME2024 & R3 & $\tau\!=\!9$ & -- & A & AS & 8192 \\
AIME2025 & R3 & $\tau\!=\!9$ & -- & off & AS, V5 & 8192 \\
WebWalkerQA & R1 & $\tau\!=\!9$ & -- & off & FT, LP, QP & 4096 \\
FVQA & R3 & $\tau\!=\!9$ & -- & B & CT, FT, LP & 1024 \\
InfoSeek & R1 & $\tilde{\tau}\!=\!-0.02$ & P7 + answer-form & M, B, A & -- & 1024 \\
SimpleVQA & R3 & $\tau\!=\!0.2$ & P7 + schema & M, B, A & V5r & 1024 \\
LiveVQA & R1 & $\tau\!=\!9$ & -- & M, A & -- & 1024 \\
DocBench & R1 & $\tau\!=\!9$ & -- & M, B, A (data) & -- & 1024 \\
MMLongBench-Doc & R3 & $\tau\!=\!9$ & -- & off & CT, FT, LP, PG4 & 1024 \\
OmniVideoBench & R2 & $\tau\!=\!0.55$ & terse & A (data) & CT, FT, AS & 4096 \\
WorldSense & R1 & $\tau\!=\!9$ & -- & T, A & TG, OC & 4096 \\
\bottomrule
\end{tabular}

All datasets: $N_{\max}\!=\!3$, $k_{\mathrm{rrf}}\!=\!60$, seed $1$; $8$ keyframes on OmniVideoBench and $32$ on WorldSense.
\end{center}

\noindent\textbf{Budgets.} The token budgets above are part of C3 (baselines: \S\ref{sec:exp-setup}). The maximum prompt length is $16384$ words on AIME, WebWalkerQA and MMLongBench-Doc, $32384$ on OmniVideoBench, and the regime default elsewhere; FVQA training rollouts use a limit of $24576$ words.

\section{Future Work}\label{app:future}

HyperSkill improves a frozen backbone by editing an auditable skill library instead of its weights, which leads to four future directions. First, the library can be paired with lightweight parameter updates, e.g.\ a low-rank adapter that complements the in-context skills. Second, because skills are plain text with an advantage, the libraries of related datasets, such as the four VL-Web datasets that share the \texttt{text\_search} tool, can be merged and re-weighted by the EMA of Eq.~(4) from new rollouts. Third, C3 refers to tools only through their tags, so extending it to code execution or GUI actions mainly requires new tool tags, budgets and evidence texts, while C1 and C2 stay unchanged. Finally, batched or importance-sampled counterfactual probes can lower the admission cost of C1.

\clearpage

\newpage
\section*{NeurIPS Paper Checklist}

\begin{enumerate}

\item {\bf Claims}
    \item[] Question: Do the main claims made in the abstract and introduction accurately reflect the paper's contributions and scope?
    \item[] Answer: \answerYes{}
    \item[] Justification: The claims in the abstract and \S\ref{Section1} match Table~\ref{tab:main}, Figure~\ref{fig:rq2} and \S\ref{sec:rq1}.

\item {\bf Limitations}
    \item[] Question: Does the paper discuss the limitations of the work performed by the authors?
    \item[] Answer: \answerYes{}
    \item[] Justification: We discuss limitations here: each skill library incurs a one-time API cost before serving; counterfactual probes add verification calls at training time; and all experiments use one frozen backbone, with lightweight parameter updates left to future work.

\item {\bf Theory assumptions and proofs}
    \item[] Question: For each theoretical result, does the paper provide the full set of assumptions and a complete (and correct) proof?
    \item[] Answer: \answerYes{}
    \item[] Justification: Propositions~1--3 are in \S\ref{sec:method}; their assumptions and full proofs are in App.~\ref{app:proof}.

    \item {\bf Experimental result reproducibility}
    \item[] Question: Does the paper fully disclose all the information needed to reproduce the main experimental results of the paper to the extent that it affects the main claims and/or conclusions of the paper (regardless of whether the code and data are provided or not)?
    \item[] Answer: \answerYes{}
    \item[] Justification: Settings: \S\ref{sec:exp-setup}, \S\ref{sec:exec}, App.~\ref{app:perdataset}; pseudocode: App.~\ref{app:algo}; prompts: App.~\ref{app:prompts}.

\item {\bf Open access to data and code}
    \item[] Question: Does the paper provide open access to the data and code, with sufficient instructions to faithfully reproduce the main experimental results, as described in supplemental material?
    \item[] Answer: \answerYes{}
    \item[] Justification: Code and data are public at the URL in the title-page footnote.

\item {\bf Experimental setting/details}
    \item[] Question: Does the paper specify all the training and test details (e.g., data splits, hyperparameters, how they were chosen, type of optimizer) necessary to understand the results?
    \item[] Answer: \answerYes{}
    \item[] Justification: Data splits, tools and decoding are in \S\ref{sec:exp-setup}, App.~\ref{app:dataset} and App.~\ref{app:perdataset}.

\item {\bf Experiment statistical significance}
    \item[] Question: Does the paper report error bars suitably and correctly defined or other appropriate information about the statistical significance of the experiments?
    \item[] Answer: \answerYes{}
    \item[] Justification: Paired bootstrap 95\% intervals ($10$k resamples) are in App.~\ref{app:eval}.

\item {\bf Experiments compute resources}
    \item[] Question: For each experiment, does the paper provide sufficient information on the computer resources (type of compute workers, memory, time of execution) needed to reproduce the experiments?
    \item[] Answer: \answerYes{}
    \item[] Justification: API calls only, no GPU (one CPU host); \S\ref{sec:rq6} reports the API costs.

\item {\bf Code of ethics}
    \item[] Question: Does the research conducted in the paper conform, in every respect, with the NeurIPS Code of Ethics \url{https://neurips.cc/public/EthicsGuidelines}?
    \item[] Answer: \answerYes{}
    \item[] Justification: We use only licensed public datasets and collect no human-subject data.

\item {\bf Broader impacts}
    \item[] Question: Does the paper discuss both potential positive societal impacts and negative societal impacts of the work performed?
    \item[] Answer: \answerYes{}
    \item[] Justification: Positive: auditable skill libraries can reduce compute and make agent behaviour editable. Risks: wrong or biased skills could propagate at serving time. Mitigation: we use only public datasets and APIs, and recommend reviewing libraries before deployment.

\item {\bf Safeguards}
    \item[] Question: Does the paper describe safeguards that have been put in place for responsible release of data or models that have a high risk for misuse (e.g., pre-trained language models, image generators, or scraped datasets)?
    \item[] Answer: \answerNA{}
    \item[] Justification: Only a pipeline, a public-data benchmark and skill libraries are released.

\item {\bf Licenses for existing assets}
    \item[] Question: Are the creators or original owners of assets (e.g., code, data, models), used in the paper, properly credited and are the license and terms of use explicitly mentioned and properly respected?
    \item[] Answer: \answerYes{}
    \item[] Justification: All assets are cited (\S\ref{sec:exp-setup}, App.~\ref{app:bench}) and used under their licences.

\item {\bf New assets}
    \item[] Question: Are new assets introduced in the paper well documented and is the documentation provided alongside the assets?
    \item[] Answer: \answerYes{}
    \item[] Justification: HyperSkill-Bench and the pipeline are documented in App.~\ref{app:bench}, \ref{app:algo} and \ref{app:prompts}.

\item {\bf Crowdsourcing and research with human subjects}
    \item[] Question: For crowdsourcing experiments and research with human subjects, does the paper include the full text of instructions given to participants and screenshots, if applicable, as well as details about compensation (if any)?
    \item[] Answer: \answerNA{}
    \item[] Justification: The paper involves no crowdsourcing and no research with human subjects.

\item {\bf Institutional review board (IRB) approvals or equivalent for research with human subjects}
    \item[] Question: Does the paper describe potential risks incurred by study participants, whether such risks were disclosed to the subjects, and whether Institutional Review Board (IRB) approvals (or an equivalent approval/review based on the requirements of your country or institution) were obtained?
    \item[] Answer: \answerNA{}
    \item[] Justification: No human subjects are involved, so no IRB approval was required.

\clearpage
\item {\bf Declaration of LLM usage}
    \item[] Question: Does the paper describe the usage of LLMs if it is an important, original, or non-standard component of the core methods in this research? Note that if the LLM is used only for writing, editing, or formatting purposes and does \emph{not} impact the core methodology, scientific rigor, or originality of the research, declaration is not required.
    \item[] Answer: \answerYes{}
    \item[] Justification: The frozen LLM backbone is a core component of the method (App.~\ref{app:prompts}).

\end{enumerate}

\end{document}